\documentclass{article}

 \usepackage{microtype}
 \usepackage{booktabs} % for professional tables

\usepackage{hyperref}
\usepackage{url}

\usepackage{hyperref}       % hyperlinks
\usepackage{url}            % simple URL typesetting
\usepackage{booktabs}       % professional-quality tables
\usepackage{amsfonts}       % blackboard math symbols
\usepackage{nicefrac}       % compact symbols for 1/2, etc.
\usepackage{microtype}      % microtypography
\usepackage{csvsimple}      % reading csv files
\usepackage{tabularx}       % equal width cells
\usepackage{xspace}         % space after commands
\usepackage{amsmath}  
\usepackage{amsopn}
\usepackage{MnSymbol}
\usepackage{graphicx}
\usepackage{subcaption}
\usepackage{makecell}
\usepackage[most]{tcolorbox}
\usepackage{multirow}
\usepackage{wrapfig}
\usepackage{bm}
\usepackage{rotating}

\newcommand{\eqq}[1]{\eqref{#1}}
\newcommand{\ud}{\mathrm{d}}
\newcommand{\R}{\mathbb{R}}
\newcommand{\N}{\mathcal{N}}
\newcommand{\I}{\mathbb{I}}

\newcommand{\latinphrase}[1]{\emph{#1}}    % italic in roman text, upshaped in italicized text
\newcommand{\ie}{\latinphrase{i.e.}\xspace}
\newcommand{\eg}{\latinphrase{e.g.}\xspace}

\newcommand{\MM}{\mathrm{MM}}
\newcommand{\sign}{\mathrm{sign}}

\newcommand{\LAM}{\texttt{LAM}\xspace}
\newcommand{\RAM}{\texttt{RAM}\xspace}

\newcommand{\refsec}[1]{Sec.~\ref{#1}}

\newcommand{\reffig}[1]{Fig.~\ref{#1}}
\newcommand{\reftab}[1]{Tab.~\ref{#1}}

\usepackage[accepted]{icml2018}

\icmltitlerunning{Transformation Autoregressive Networks}

\begin{document}

\twocolumn[
\icmltitle{Transformation Autoregressive Networks}

% It is OKAY to include author information, even for blind
% submissions: the style file will automatically remove it for you
% unless you've provided the [accepted] option to the icml2018
% package.

% List of affiliations: The first argument should be a (short)
% identifier you will use later to specify author affiliations
% Academic affiliations should list Department, University, City, Region, Country
% Industry affiliations should list Company, City, Region, Country

% You can specify symbols, otherwise they are numbered in order.
% Ideally, you should not use this facility. Affiliations will be numbered
% in order of appearance and this is the preferred way.
\icmlsetsymbol{equal}{*}

\begin{icmlauthorlist}
\icmlauthor{Junier B Oliva}{unc,cmu} %\phantom{space}
\icmlauthor{Avinava Dubey}{cmu} %\phantom{space} 
\icmlauthor{Manzil Zaheer}{cmu}

\icmlauthor{Barnab{\'a}s P{\'o}czos}{cmu} %\phantom{space} 
\icmlauthor{Ruslan Salakhutdinov}{cmu} %\phantom{space} 
\icmlauthor{Eric P Xing}{cmu} %\phantom{space}
\icmlauthor{Jeff Schneider}{cmu}
%\icmlauthor{Machine Learning Department, Carnegie Mellon University, Pittsburgh, PA 15213}{cmu}
\end{icmlauthorlist}

\icmlaffiliation{cmu}{Machine Learning Department, Carnegie Mellon University, Pittsburgh, PA 15213}
\icmlaffiliation{unc}{Computer Science Department, University of North Carolina, Chapel Hill, NC 27599 (Work completed while at CMU.)}
% \icmlaffiliation{goo}{Googol ShallowMind, New London, Michigan, USA}
% \icmlaffiliation{ed}{School of Computation, University of Edenborrow, Edenborrow, United Kingdom}

\icmlcorrespondingauthor{Junier Oliva}{\mbox{joliva@cs.unc.edu}}
%\icmlcorrespondingauthor{}{ep@eden.co.uk}

% You may provide any keywords that you
% find helpful for describing your paper; these are used to populate
% the "keywords" metadata in the PDF but will not be shown in the document
\icmlkeywords{Density Estimation, Autoregressive Models, RNNs}

\vskip 0.3in
]

% this must go after the closing bracket ] following \twocolumn[ ...

% This command actually creates the footnote in the first column
% listing the affiliations and the copyright notice.
% The command takes one argument, which is text to display at the start of the footnote.
% The \icmlEqualContribution command is standard text for equal contribution.
% Remove it (just {}) if you do not need this facility.

\printAffiliationsAndNotice{}  % leave blank if no need to mention equal contribution
%\printAffiliationsAndNotice{\icmlEqualContribution} % otherwise use the standard text.

\begin{abstract}
The fundamental task of general density estimation $p(x)$ has been of keen interest to machine learning. 
%Recent advances in density estimation have been fragmented and myopic in their study. 
In this work, we attempt to systematically characterize methods for density estimation.  Broadly speaking, most of the existing methods can be categorized into either using: \textit{a}) autoregressive models to estimate the conditional factors of the chain rule, $p(x_{i}\, |\, x_{i-1}, \ldots)$; or \textit{b}) non-linear transformations of variables of a simple base distribution. 
%The systematic study revealed that some straightforward, yet novel approaches and combinations thereof were missed earlier. 
%For example, the untied linear conditionals that we found in our experimental evaluations to work better.
Based on the study of the characteristics of these categories, we propose multiple novel methods for each category. For example we propose RNN based transformations to model non-Markovian dependencies.
%Instead, this work jointly leverages transformations of variables and autoregressive conditional models, and proposes novel methods for both.
Further, through a comprehensive study over both real world and synthetic data, we show that jointly leveraging transformations of variables and autoregressive conditional models, results in a considerable improvement in performance. We illustrate the use of our models in outlier detection and image modeling. Finally we introduce a novel data driven framework for learning a family of distributions.
% \\
% \\
% The fundamental task of general density estimation has been of keen interest to machine learning. Recent advances in density estimation have either: \textit{a}) proposed a flexible model to estimate the conditional factors of the chain rule, $p(x_{i}\, |\, x_{i-1}, \ldots)$; or \textit{b}) used flexible, non-linear transformations of variables of a simple base distribution. 
% Instead, this work jointly leverages transformations of variables and autoregressive conditional models, and proposes novel methods for both. We provide a deeper understanding of our methods, showing a considerable improvement through a comprehensive study over both real world and synthetic data. Moreover, we illustrate the use of our models in outlier detection and image modeling tasks.
\end{abstract}

%\vspace{-2mm}
\section{Introduction}
Density estimation is at the core of a multitude of machine learning applications. 
However, this fundamental task
%, which encapsulates the understanding of data, 
is difficult in the general setting due to issues like the curse of dimensionality. Furthermore, for general data, unlike spatial/temporal data, we do not have known correlations a priori among covariates that may be exploited. For example, image data has known correlations among neighboring pixels that may be hard-coded into a model through convolutions, whereas one must find such correlations in a data-driven fashion with general data.

\begin{figure}[t]
\centering
\includegraphics[width=\columnwidth, trim={0 10mm 0mm 3mm}, clip]{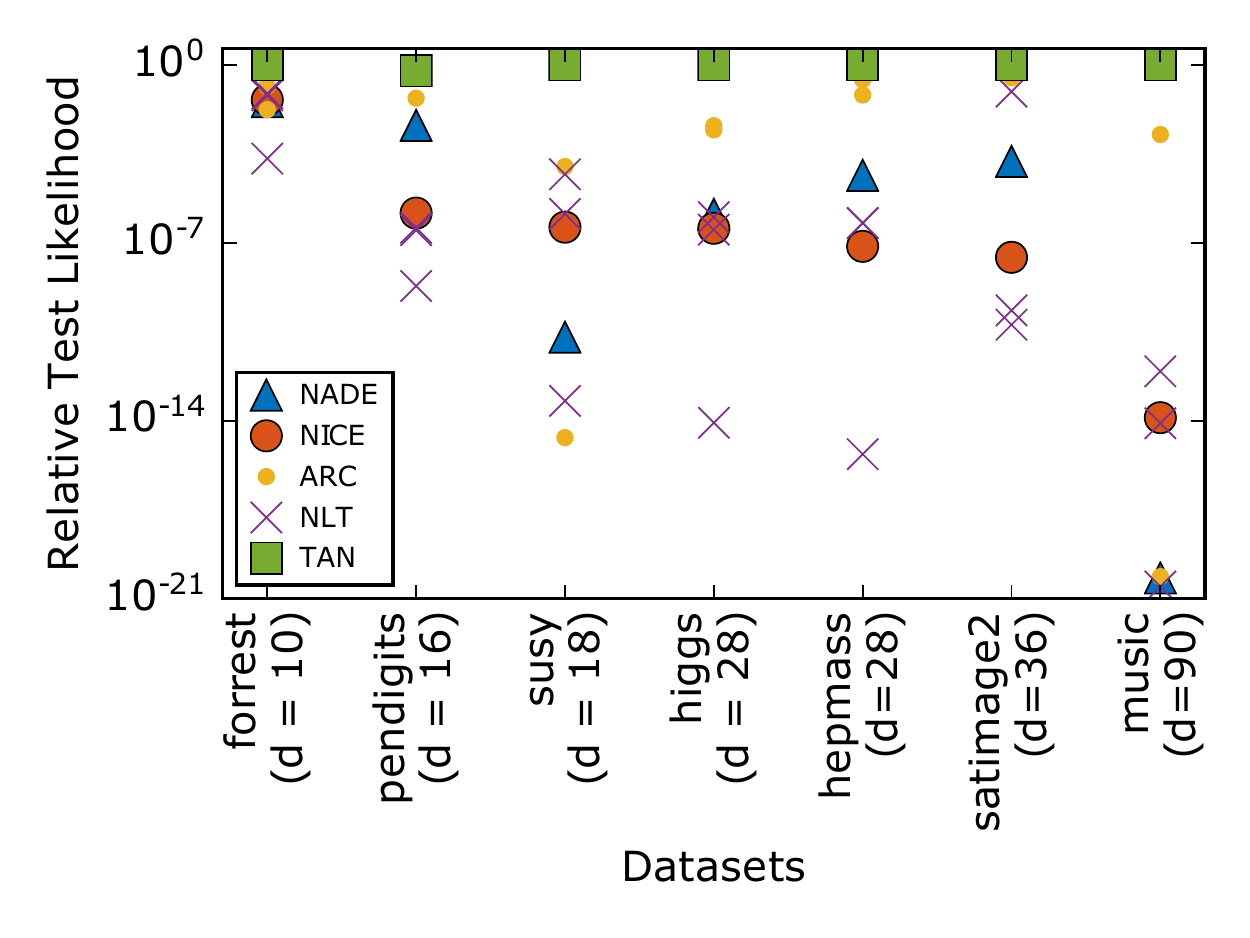}
\vspace{-8mm}
\caption{The proposed \textit{TAN} models for density estimation, which jointly leverages non-linear transformation and autoregressive conditionals, shows considerable improvement over other methods across datasets of varying dimensions. The scatter plots shows that only utilizing autoregressive conditionals (ARC) without transformations (\eg existing works like NADE \citep{uria1} and other variants) or only using non-linear transformation (NLT) with simple restricted conditionals (\eg existing works like NICE \citep{dinh1} and other variants) is not sufficient for all datasets.\label{fig:teaser}}
\vspace{-4mm}
\end{figure}

In order to model high dimensional data, the main challenge lies in constructing models that are flexible enough while having tractable learning algorithms. A variety of diverse solutions exploiting different aspects of the problems have been proposed in the literature.  A large number of methods have considered auto-regressive models to estimate the conditional factors $p(x_i | x_{i-1}, \ldots, x_{1})$, for $i \in \{1,\ldots,d\}$ in the chain rule  \citep{larochelle2011neural, uriaRNADE, uria2, germain2015made, gregor2014deep}.
%These models estimate the conditionals:.
While some methods directly model the conditionals $p(x_i | x_{i-1}, \ldots)$ using sophisticated semiparametric density estimates, other methods apply sophisticated transformations of variables $x \mapsto z$ and take the conditionals over $z$ to be a restricted, often independent base distribution $p(z_i | z_{i-1}, \ldots) \approx f(z_i)$ \citep{dinh1, dinh2}. %, goodfellow2016nips}.
Further related works are discussed in \refsec{sec:related}.
However, looking across a diverse set of dataset, as in \reffig{fig:teaser},  neither of these approaches have the flexibility required to accurately model real world data. 
%In this paper we take a step back to first organize 
% from these previous approaches that have considered either: \textit{a}) a flexible autoregressive scheme with simple or no transformations of variables; or \textit{b}) a simple autoregressive scheme with flexible transformations of variables.
%We leverage both of these approaches (Figure~\ref{fig:flext_flexc}), develop novel methods for each, and show a considerable improvement with their combination.

%\paragraph{Contributions:}
%The following are our contributions. 
In this paper we take a step back and start from the basics. 
%The process of density estimation consists of two operations: \textit{a}) transformation of the variables and \textit{b}) modeling the conditionals. 
If we only model the conditionals, the conditional factors $p(x_i | x_{i-1}, \ldots)$, may become increasingly complicated as $i$ increases to $d$. On the other hand if we use a complex transformation with restricted conditionals then the transformation has to ensure that the transformed variables are independent. 
%In the absence of transformation the conditional factors may have to capture complex correlations between covariates. Moreover in the absence of sophisticated estimates of the conditional, the transformation has to ensure that the transformed variables are independent.  
This requirement of independence on the transformed variables can be very restrictive. Now note that the transformed space is homeomorphic to the original space and a simple relationship between the density of the two spaces exists through the Jacobian. Thus, we can employ conditional modeling on the transformed variables to alleviate the independence requirement, while being able to recover density in the original space in a straightforward fashion. In other words, we propose \textit{transformation autoregressive networks} (TANs) which composes the complex transformations and autoregressive modeling of the conditionals.
%
%A continuous bijective transformation $x \mapsto z$ having a tractable Jacobian can be composed with the complex conditional factors of the transformed variable $p(z_i | z_{i-1}, \ldots)$ to give a resulting density estimation for $p(x)$.
%Moreover, the two components can compose with each other.
%
The composition not only increases the flexibility of the model but also reduces the expressibility power needed from each of the individual components. This leads to an improved performance as can be seen from \reffig{fig:teaser}.

%A composite of transformations coupled with the LAM/RAM networks provides a highly expressive model for modeling arbitrary joint densities but retaining interpretable conditional structure.

In particular, first we propose two flexible autoregressive models for modeling conditional distributions: the linear autoregressive model (LAM), and the recurrent autoregressive model (RAM) (\refsec{sec:autoreg}).
%LAM employs a simple linear form to condition on previously seen covariates in a flexible fashion. RAM uses a recurrent neural network (RNN) to evolve conditioning features as the set of conditioning covariates expands
%To the best of our knowledge, this is the first application of RNNs for estimating conditionals in general non-spatial/temporal data. 
Secondly, we introduce several novel transformations of variables: 1) an efficient method for learning a linear transformation on covariates; 2) an invertible RNN-based transformation that directly acts on covariates; 3) an additive RNN-base transformation (\refsec{sec:tanTrans}).
%To better capture correlations in general data, we propose transformation autogressive networks (TANs) that combine our novel autoreggresive models and transformations of variables. %That is, we propose transformation autogressive networks (TANs) that use a flexible transformation over the original covariates and a powerful autoregressive model to estimate the conditional probabilities over the transformed space.
%To better capture correlations in general data, we combine our novel autoreggresive models and transformations of variables. That is, we propose transformation autogressive networks (TANs) that use a flexible transformation over the original covariates and a powerful autoregressive model to estimate the conditional probabilities over the transformed space.
%To assess the efficacy of models, we performed a comprehensive evaluation of autoregressive models and transformations. 
%We performed a comprehensive evaluation of autoregressive models and transformations that shows the fundamental result that modern density estimation methods should employ \emph{both} a flexible conditional model and a flexible transformation.
Extensive experiments on both synthetic (\refsec{sec:exptSynth}) and real-world (\refsec{sec:exptReal}) datasets show the power of TANs for capturing complex dependencies between the covariates. We run an ablation study to demonstrate contributions of various components in TAN \refsec{sec:exptAbl},
%Experiments considered both synthetic and real world datasets, showing held-out test likelihoods 
Moreover, we show that the learned model can be used for anomaly detection (\refsec{sec:exptAnom}) and learning a family of distributions (\refsec{sec:exptEmbd}).

%\input{figures/TANfigure}

% The remainder of the paper is structured as follows. 
% %First, we describe our models in detail. 
% First, in Section~\ref{sec:autoreg} we present two novel methods for modeling condition distributions across covariates. Next, in Section~\ref{sec:Trans}, we describe several transformations to use in conjunction with our proposed conditional models. After, we discuss related work and contrast our approach to previous methods. We then illustrate the efficacy of our methods with both synthetic and real-world data.

\section{Transformation Autoregressive Networks}
As mentioned above, TANs are composed of two modules: \textit{a}) an autoregressive module for modeling conditional factors and \textit{b}) transformations of variables. We first introduce our two proposed 
%novel % TODO: other word?
autoregressive models to estimate the conditional distribution of input covariates $x\in\R^d$. Later, we  show how to use such models over transformation $z=q(x)$, while renormalizing to obtain density values for $x$.
\vspace{-2mm}
\subsection{Autoregressive Models\label{sec:autoreg}}
\vspace{-2mm}
Autoregressive models decompose density estimation of a multivariate variable $x\in \R^d$ into multiple conditional tasks on a growing set of inputs through the chain rule:
\vspace{-2mm}
\begin{equation}
p(x_1, \ldots, x_d) = \prod_{i=1}^d p(x_i\, |\, x_{i-1}, \ldots, x_{1}). \label{eq:chain_rule}
\vspace{-1.5mm}
\end{equation}
That is, autoregressive models will estimate the $d$ conditional distributions $p(x_i | x_{i-1}, \ldots)$.
A class of autoregressive models can be defined by approximating conditional distributions through a mixture model, $\MM(\theta(x_{i-1}, \ldots, x_{1}))$, with parameters $\theta$ depending on $x_{i-1}, \ldots, x_{1}$:
\vspace{-1.5mm}
\begin{align}
    p(x_i | x_{i-1}, \ldots, x_{1}) &= p(x_i\, |\, \MM(\theta(x_{i-1}, \ldots, x_{1})), \label{eq:auto_MM}\\
    \theta(x_{i-1}, \ldots, x_{1}) &= f\left(h_{i}\right)  \\
    h_{i} &= g_{i}\left(x_{i-1}, \ldots, x_{1}\right), \label{eq:auto_hidden}
\vspace{-2mm}
\end{align}
where $f(\cdot)$ is a fully connected network that may use a element-wise non-linearity on inputs, and $g_i(\cdot)$ is some general mapping that computes a hidden state of features, $h_i\in\R^p$, which help in modeling the conditional distribution of $x_{i}\,|\,x_{i-1}, \ldots, x_{1}$. One can control the flexibility of the model through $g_i$. It is important to be powerful enough to model our covariates while still generalizing. In order to achieve this we propose two methods for modeling $g_i$.

\vspace{-3mm}
\paragraph{Linear Autoregressive Model (LAM):}  
This uses a straightforward linear map as $g_{i}$ in \eqref{eq:auto_hidden}:
\vspace{-2mm}
\begin{equation}
g_{i}\left(x_{i-1}, \ldots, x_{1}\right) = W^{(i)}x_{<i} + b \label{eq:untied},
\vspace{-1mm}
\end{equation}
where $W^{(i)}\in \R^{p \times (i-1)}$, $b \in \R^{p}$, and $x_{<i} = (x_{i-1}, \ldots, x_{1})^T$. 
Notwithstanding the simple form of \eqq{eq:untied}, the resulting model is quite flexible as it may model consecutive conditional problems $p(x_i | x_{i-1}, \ldots, x_{1})$ and $p(x_{i+1} | x_{i}, \ldots, x_{1})$ very differently owing to different $W^{(i)}$s.

\vspace{-3mm}
\paragraph{Recurrent Autoregressive Model (RAM):}
This features a recurrent relation between $g_i$'s.
As the set of covariates is progressively fed into $g_i$'s, it is natural to consider a hidden state evolving according to an RNN recurrence relationship:
%Given the expanding set of covariates progressively fed into $g_i$'s, it is natural to consider a hidden state that evolves according to an RNN recurrence relationship:
\vspace{-2mm}
\begin{equation}
    h_{i} = g\left(x_{i-1}, g(x_{i-2}, \ldots, x_{1})\right) = 
    g\left(x_{i-1}, h_{i-1}\right). \label{eq:rnn_model}
\vspace{-1mm}
\end{equation}
In this case $g(x, s)$ is a RNN function for updating one's state based on an input $x$ and prior state $s$. In the case of gated-RNNs, the model will be able to scan through previously seen dimensions remembering and forgetting information as needed for conditional densities without making any strong Markovian assumptions.

Both LAM and RAM are flexible and able to adjust the hidden states, $h_i$ in \eqref{eq:auto_hidden}, to model the distinct conditional tasks $p(x_i | x_{i-1}, \ldots)$. There is a trade-off of added flexibility and transferred information between the two models. LAM treats the conditional tasks for $p(x_i | x_{i-1}, \ldots)$ and $p(x_{i+1} | x_{i}, \ldots)$  in a largely independent fashion. This makes for a very flexible model, however the parameter size is also large 
and there is no sharing of information among the conditional tasks. 
On the other hand, RAM provides a framework for transfer learning among the conditional tasks by allowing the hidden state $h_{i}$ to evolve 
through the distinct
%as we consider different 
conditional tasks. This leads to fewer parameters 
and more sharing of information in respective tasks, but also yields less flexibility since conditional estimates are tied, and may only change in a smooth fashion.

\vspace{-2mm}
\subsection{Transformations\label{sec:tanTrans}}
\vspace{-2mm}
%Several methods \citep{dinh1, dinh2, goodfellow2016nips} have shown that the expressive power of very simple conditional densities \eqq{eq:chain_rule} (such as independent Gaussians) can be greatly improved with transformations of variables. Although the chain rule holds for arbitrary distributions, a limited amount of data and parameters limits the expressive power of models. Hence, we expect that combining our conditional models with transformations of variables will also further increase flexibility. 
Next we introduce the second module of TANs i.e. the transformations.
When using an invertible transformation of variables $z=(q_1(x), \ldots, q_d(x)) \in \R^d$, one can establish a relationship between the pdf of $x$ and $z$ as:
\vspace{-2mm}
\begin{equation}
p(x_1, \ldots, x_d) = \left|\det\frac{\ud q}{\ud x}\right|\, \prod_{i=1}^d p\left(z_i\, |\, z_{i-1}, \ldots, z_1 \right), \label{eq:detloglike}
\vspace{-2mm}
\end{equation}
where $\left|\det\frac{\ud q}{\ud x}\right|$ is the Jacobian of the transformation. 
For analytical and computational considerations, we require transformations to be invertible, efficient to compute and invert, and have a structured Jacobian matrix. In order to meet these criteria we propose the following transformations. 

\vspace{-3mm}
\paragraph{Linear Transformation:}
It is an affine map of the form:
\vspace{-2mm}
\begin{equation}
z = A x + b, \label{eq:linear_trans}
\vspace{-2mm}
\end{equation}
where we take $A$ to be invertible. Note that even though this linear transformation is simple, it includes permutations, and may also perform a PCA-like transformation, capturing coarse and highly varied features of the data before moving to more fine grained details. In order to not incur a high cost for updates, we wish to compute the determinant of the Jacobian efficiently. Thus, we propose to directly work over an LU decomposition $A = LU$ where $L$ is a lower triangular matrix with unit diagonals and $U$ is a upper triangular matrix with arbitrary diagonals. As a function of $L$, $U$ we have that $\det\frac{\ud z}{\ud x} = \prod_{i=1}^d U_{ii}$; hence we may efficiently optimize the parameters of the linear map. Furthermore, inverting our mapping is also efficient through solving two triangular matrix equations.

\vspace{-3mm}
\paragraph{Recurrent Transformation:\label{sec:tanRT}} Recurrent neural networks are also a natural choice for variable transformations. Due to their dependence on only previously seen dimensions, RNN transformations have triangular Jacobians, leading to simple determinants. Furthermore, with an invertible output unit, their inversion is also straight-forward. We consider the following form to an RNN transformation:
\vspace{-2mm}
{
\begin{equation}
\small
\begin{aligned}
z_i = r_\alpha\left( y x_i + w^{T} s_{i-1} + b \right),\  
s_i = r\left( u x_i + v^{T} s_{i-1} + a \right), \label{eq:rnn_trans}
\end{aligned}
\vspace{-2mm}
\end{equation}
}
\normalsize
where $r_\alpha$ is a leaky ReLU unit $r_\alpha(t) = \I\{t<0\} \alpha t + \I\{t\ge0\} t$, $r$ is a standard ReLU unit, $s\in \R^{\rho}$ is the hidden state  $y$, $u$, $b$, $a$ are scalars, and $w, v \in \R^{\rho}$ are vectors.
As compared to the linear transformation, the recurrent transformation is able to transform the input with different dynamics depending on its values.
Inverting \eqref{eq:rnn_trans} is a matter of inverting outputs and updating the hidden state (where the initial state $s_0$ is known and constant):
\vspace{-2mm}
\begin{equation}
\small
\begin{aligned}
 x_i &= \frac{1}{y}\left(r^{-1}_\alpha\left(z^{(r)}_i\right) - w^{T} s_{i-1} - b\right), \\
 s_i &= r\left( u x_i + v^{T} s_{i-1} + a \right). \label{eq:rnn_trans_inv}
\end{aligned}
\vspace{-2mm}
\end{equation}
\normalsize
Furthermore, the determinant of the Jacobian for \eqq{eq:rnn_trans} is the product of diagonal terms:
\vspace{-2mm}
\begin{equation}
    \det\frac{\ud z}{\ud x} = y^d \prod_{i=1}^d  r^{\prime}_\alpha\left( y x_i + w^{T} s_{i-1} + b \right)
    \label{eq:rnn_det},
\vspace{-2mm}
\end{equation}
where $r^{\prime}_\alpha\left(t\right) = \I\{t> 0\} + \alpha \I\{t<0\}$. 
%For added dependence among all dimensions, we consider a forward pass of a recurrent transformation \eqq{eq:rnn_trans} followed by another recurrent transformation in a backwards pass. 

%In total we consider three transformations to covariates, an initial linear mapping, then forwards and backwards passes of recurrent transformations: $x \mapsto z^{(\ell)} \mapsto z^{(f)} \mapsto z^{(b)}$
%(see Figure~\ref{fig:transformation}).

%Lastly, we note that sampling a RED model is computationally efficient. One must simply propagate random draws of dimensions through the conditional RNN: $\ddot{z}^{(b)}_{k} \sim p({z}^{(b)} | \ddot{z}^{(b)}_{k-1}, \ldots, \ddot{z}^{(b)}_{1}) = p({z}^{(b)} | \ddot{h}_{k-1})$, where $\ddot{z}^{(b)}_{j}$ are the drawn dimensions of the latent space and $\ddot{h}_{k-1}$ is the corresponding conditional hidden state after seeing $\ddot{z}^{(b)}_{k-1}, \ldots, \ddot{z}^{(b)}_{1}$. After, one inverts the linear/RNN transformation to produce a sample, $\ddot{x} = q^{-1}(\ddot{z}^{(b)})$. As explained above, inverting this transformation is inexpensive, especially when caching $A^{-1}$ (a one time cost of inverting the product of two triangular matrices $L, U$).

\vspace{-3mm}
\paragraph{Recurrent Shift Transformation: \label{sec:tanrst}} It is worth noting that the rescaling brought on by the recurrent transformation effectively incurs a penalty through the log of the determinant \eqq{eq:rnn_det}. However, one can still perform a transformation that depends on the values of covariates through a shift operation. In particular, we propose an additive shift based on a recurrent function on prior dimensions:
\vspace{-1.5mm}
\begin{equation}
    z_i = x_i + m(s_{i-1}),\quad
    s_i = g(x_i, s_{i-1}), \label{eq:rnn_shift}
    \vspace{-2mm}
\end{equation}
where $g$ is recurrent function for updating states, and $m$ is a fully connected network. Inversion proceeds as before:
\vspace{-1.5mm}
\begin{equation}
    x_i = z_i - m(s_{i-1}),\quad
    s_i = g(x_i, s_{i-1}). \label{eq:rnn_shift_inv}
    \vspace{-2mm}
\end{equation}
The Jacobian is again lower triangular, however due to the additive nature of \eqq{eq:rnn_shift}, we have a unit diagonal. Thus, 
$\det\frac{\ud z}{\ud x} = 1$. One interpretation of this transformation is that one can shift the value of $x_k$ based on $x_{k-1}, x_{k-2}, \ldots$ for better conditional density estimation without any penalty coming from the determinant term in \eqq{eq:detloglike}.

%Notwithstanding their natural fit, this is, to the best of our knowledge, the first use of recurrent networks for transformations of variables in general data density estimation.

% \paragraph{Additive Coupling Transformation:} We also consider the use of the additive coupling transformation introduced in \citep{dinh1}. Additive coupling proceeds by splitting inputs into halves $x = ( x_{< {d}/{2}},\, x_{\geq {d}/{2}} )$, and transforming the second half as an additive function of the first half:
% \begin{align}
%   z = \left( x_{< {d}/{2}},\, x_{\geq {d}/{2}} + m(x_{< {d}/{2}}) \right) \label{eq:add_coup},
% \end{align}
% where $m(\cdot)$ is the output of a fully connected network.
% Inversion is simply a matter of subtraction:
% \begin{align}
%   x = \left( z_{< {d}/{2}},\, z_{\geq {d}/{2}} + m(z_{< {d}/{2}}) \right) \label{eq:add_coup_inv}.
% \end{align}
% Furthermore, as with the RNN shift transformation, the additive nature of \eqq{eq:add_coup} yeilds a simple determinant, $\det\frac{\ud z}{\ud x} = 1$.

\vspace{-3mm}
\paragraph{Composing Transformations:} Lastly, we considering stacking (i.e. composing) several transformations
$q = q^{(1)} \circ \ldots \circ q^{(T)}$ and renormalizing:
\vspace{-2mm}
\begin{equation}
\begin{aligned}
p(x_1, \ldots, x_d) = &\prod_{t=1}^T\left|\det\frac{\ud q^{(t)}}{\ud q^{(t-1)}}\right| \\
 &\prod_{i=1}^d p\left(q_i(x)\, |\, q_{i-1}(x), \ldots, q_1(x)\right) \label{eq:multitrans},
\end{aligned}
\vspace{-2mm}
\end{equation}
where we take $q^{(0)}$ to be $x$.
We note that composing several transformations together allows one to leverage the respective strengths of each transformation. Moreover, inserting a reversal mapping ($x_1,\ldots,x_d \mapsto x_d,\ldots,x_1$) as one of the $q_i$s yields bidirectional relationships.

\vspace{-2mm}
\subsection{Combined Approach}
\vspace{-2mm}
We combine the use of both transformations of variables and rich autoregressive models by: 1) writing the density of inputs, $p(x)$, as a normalized density of a transformation: $p(q(x))$ \eqq{eq:multitrans}. Then we estimate the conditionals of $p(q(x))$ using an autoregressive model, \ie, to learn our model we minimize the negative log likelihood: 
\vspace{-2mm}
\small
\begin{equation}
\begin{aligned}
    &- \log p(x_1, \ldots, x_d) =\\
    &- \sum_{t=1}^T\log\left|\det\frac{\ud q^{(t)}}{\ud q^{(t-1)}}\right| 
    -\sum_{i=1}^d \log p\left(q_i(x)\, |\, h_{i}\right),
    \label{eq:nll}
\end{aligned}
\vspace{-2mm}
\end{equation}
\normalsize
which is obtained by substituting \eqq{eq:auto_MM} into \eqq{eq:multitrans} with $h_i$ as defined in \eqq{eq:auto_hidden}.

\section{Related Works} \label{sec:related}
Nonparametric density estimation has been a well studied problem in statistics and machine learning \citep{larry}. Unfortunately, nonparametric approaches like kernel density estimation suffer greatly from the curse of dimensionality and do not perform well when data does not have a small number of dimensions ($d \lesssim 3$). To alleviate this, several semiparametric approaches have been explored. Such approaches include forest density estimation \citep{lafferty}, which assumes that the data has a forest (i.e. a collection of trees) structured graph. This assumption leads to a density which factorizes in a first order Markovian fashion through a tree traversal of the graph. Another common semiparametric approach is to use a nonparanormal type model \citep{liu}. This approach uses a Gaussian copula with a rank-based transformation and a sparse precision matrix. While both approaches are well-understood theoretically, their strong assumptions lead to inflexible models.

In order to provide greater flexibility with semiparametric models, recent work has employed deep learning for density estimation. The use of neural networks for density estimation dates back to \citet{bishop} and has seen success in speech \citep{speech, Uriaspeech}, music \citep{NicolasMusic}, etc. Typically such approaches use a network to learn the parameters of a parametric model for data. Recent work has also explored the application of deep learning to build density estimates in image data \citep{pixel, dinh2}. However, such approaches are heavily reliant on exploiting structure in neighboring pixels, often subsampling, reshaping or re-ordering data, and using convolutions to take advantage of neighboring correlations. Modern approaches for general density estimation in real-valued data include \citet{uriaRNADE, uria2,germain2015made,gregor2014deep,dinh1, kingma2016improving, papamakarios2017masked}.

NADE \citep{uriaRNADE} is an RBM-inspired density estimator with a weight-sharing scheme across conditional densities on covariates. It may be written as a special case of LAM \eqref{eq:untied} with tied weights:
\vspace{-2mm}
\begin{equation}
q_{i}\left(x_{i-1}, \ldots, x_{1}\right) = W_{<i}x_{<i} + b, \label{eq:nade}
\vspace{-2mm}
\end{equation}
where $W_{<i} \in \R^{p \times {i-1}}$ is the weight matrix compose of the first $i-1$ columns of a shared matrix $W = (w_1, \ldots w_d)$.
We note also that LAM and NADE are both related to fully visible sigmoid belief networks \citep{frey1998graphical, neal1992connectionist}.

% TODO: more rigorous way of say this?
Even though the weight-sharing scheme in \eqref{eq:nade} reduces the number of parameters, it also limits the types of distributions one can model. Roughly speaking, the NADE weight-sharing scheme makes it difficult to adjust conditional distributions when expanding the conditioning set with a covariate that has a small information gain. 
We illustrate this
%these kinds of limitations 
by considering a simple 3-dimensional distribution:
%\begin{align}
    $x_1 \sim \N(0, 1)$, %\quad
    $x_2 \sim \N(\sign(x_1), \epsilon)$, %\quad
    $x_3 \sim \N\left(\I\left\{|x_1| < C_{0.5}\right\}, \epsilon\right)$, %\label{eq:dist_example}
%\end{align}
where $C_{0.5}$ is the $50\%$ confidence interval of a standard Gaussian distribution, and $\epsilon>0$ is some small constant. That is, $x_2$, and $x_3$ are marginally distributed as an equi-weighted bimodal mixture of Gaussian with means $-1, 1$ and $0, 1$, respectively. 
%Using, \eqq{eq:nade} we see that the states determining the distributions of $x_2$ and $x_3$ are $h_2 = w_1 x_1 + b$ and $h_3 = w_1 x_1 + w_2 x_2 + b$, respectively. That is, the difference in the distribution of $x_3$ from that of $x_2$ must be captured entirely by $w_2 x_2$. However, the sign of $x_1$ will not be informative of its magnitude, hence weight-sharing will prohibit one from correctly modeling $x_3$, notwithstanding the fact that it is conditioned on $x_1$. I.e. the conditional model for $x_2$ will suffer if we try to model $x_3$ better and vice-versa.
Due to NADE's weight-sharing linear model, it will be difficult to adjust $h_2$ and $h_3$ jointly to correctly model $x_2$ and $x_3$ respectively.
However, given their additional flexibility, both LAM and RAM are able to adjust hidden states to remember and transform features as needed.

NICE \citep{dinh1} and its successor Real NVP \citep{dinh2}  models assume that data is drawn from a latent independent Gaussian space and transformed. The transformation uses several ``additive coupling'' shifting on the second half of dimensions, using the first half of dimensions. For example NICE's
additive coupling proceeds by splitting inputs into halves $x = ( x_{< {d}/{2}},\, x_{\geq {d}/{2}} )$, and transforming the second half as an additive function of the first half:
\vspace{-2mm}
\begin{equation}
   z = \left( x_{< {d}/{2}},\, x_{\geq {d}/{2}} + m(x_{< {d}/{2}}) \right) \label{eq:add_coup},
\vspace{-2mm}
\end{equation}
where $m(\cdot)$ is the output of a fully connected network.
Inversion is simply a matter of subtraction $
   x = \left( z_{< {d}/{2}},\, z_{\geq {d}/{2}} - m(z_{< {d}/{2}}) \right) \label{eq:add_coup_inv}.
$
The full transformation is the result of stacking several of these additive coupling layers together followed by a final rescaling operation. Furthermore, as with the RNN shift transformation, the additive nature of \eqq{eq:add_coup} yields a simple determinant, $\det\frac{\ud z}{\ud x} = 1$.

MAF \citep{papamakarios2017masked} identified that Gaussian conditional autoregressive models for density estimation can be seen as transformations. This enabled them to stack multiple autoregressive models that increases flexibility. However, stacking Gaussian conditional autoregressive models amounts to just stacking shift and scale transformations. Unlike MAF, in the TAN framework we not only propose novel and more complex equivalence like Recurrent Transformation (\refsec{sec:tanTrans}), but also systematically composing stacks of such transformations with flexible autoregressive models.

%We note that 
There are several methods for obtaining samples from an unknown distribution that by-pass density estimation. For instance, generative adversarial networks (GANs) apply a (typically noninvertible) transformation of variables to a base distribution by optimizing a minimax loss 
%over a discrimator and the transformation 
\citep{goodfellow2016nips,kingma2016improving}. Samples can also be obtain from methods that compose graphical models with deep networks \citep{johnson2016composing,al2017contextual}. 
Furthermore, one can also obtain samples with only limited information about the density of interest using methods such as Markov chain Monte Carlo \citep{neal1993probabilistic}, Hamiltonian Monte Carlo \citep{Neal2010MCMC}, stochastic variants \citep{dubey2016variance}, etc. 

\section{Experiments}
We now present empirical studies for our TAN framework in order to establish (i) the superiority of TANs over one-prong approaches (\refsec{sec:exptSynth}), (ii) that TANs are accurate on real world datasets (\refsec{sec:exptReal}), (iii) the importance of various components of TANs, (iv) that TANs are easily amenable to various tasks (\refsec{sec:exptAnom}), such as learning a parametric family of distributions and being able to generalize over unseen parameter values (\refsec{sec:exptEmbd}).

% We compare  models using several experiments on synthetic and real-world datasets. First, we compute the average log likelihoods on test data. Then, to gain further context of the efficacy of models, we also use their density estimates for anomaly detection, where we take low density instances to be outliers. Moreover, we look at an illustrative MNIST image modeling task. 

\begin{figure*}[t]
    \centering
    \vspace{-1mm}
    \begin{subfigure}[b]{0.24\textwidth}
        \includegraphics[width=\textwidth]{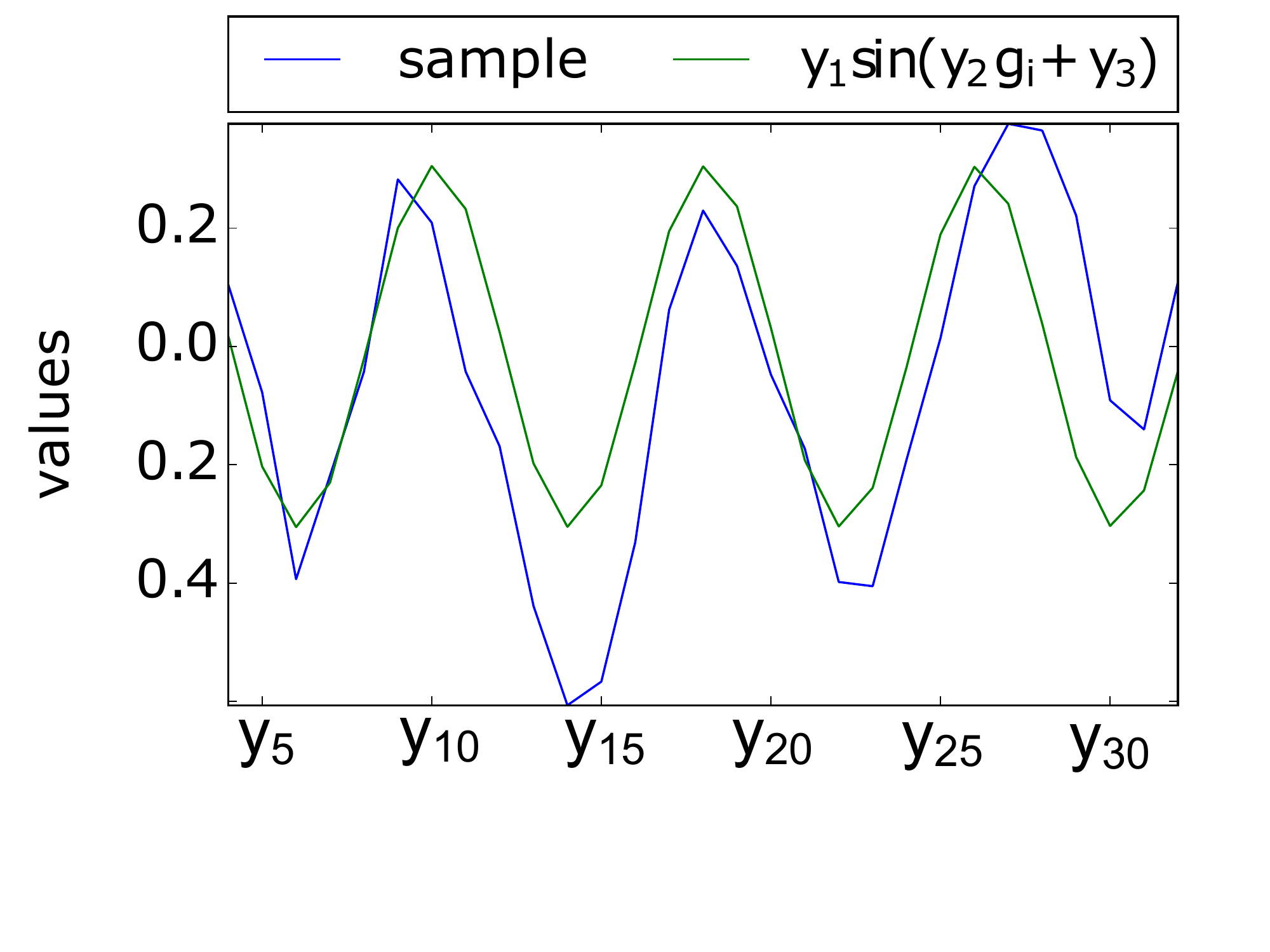}
    \end{subfigure}
    \begin{subfigure}[b]{0.24\textwidth}
        \includegraphics[width=\textwidth]{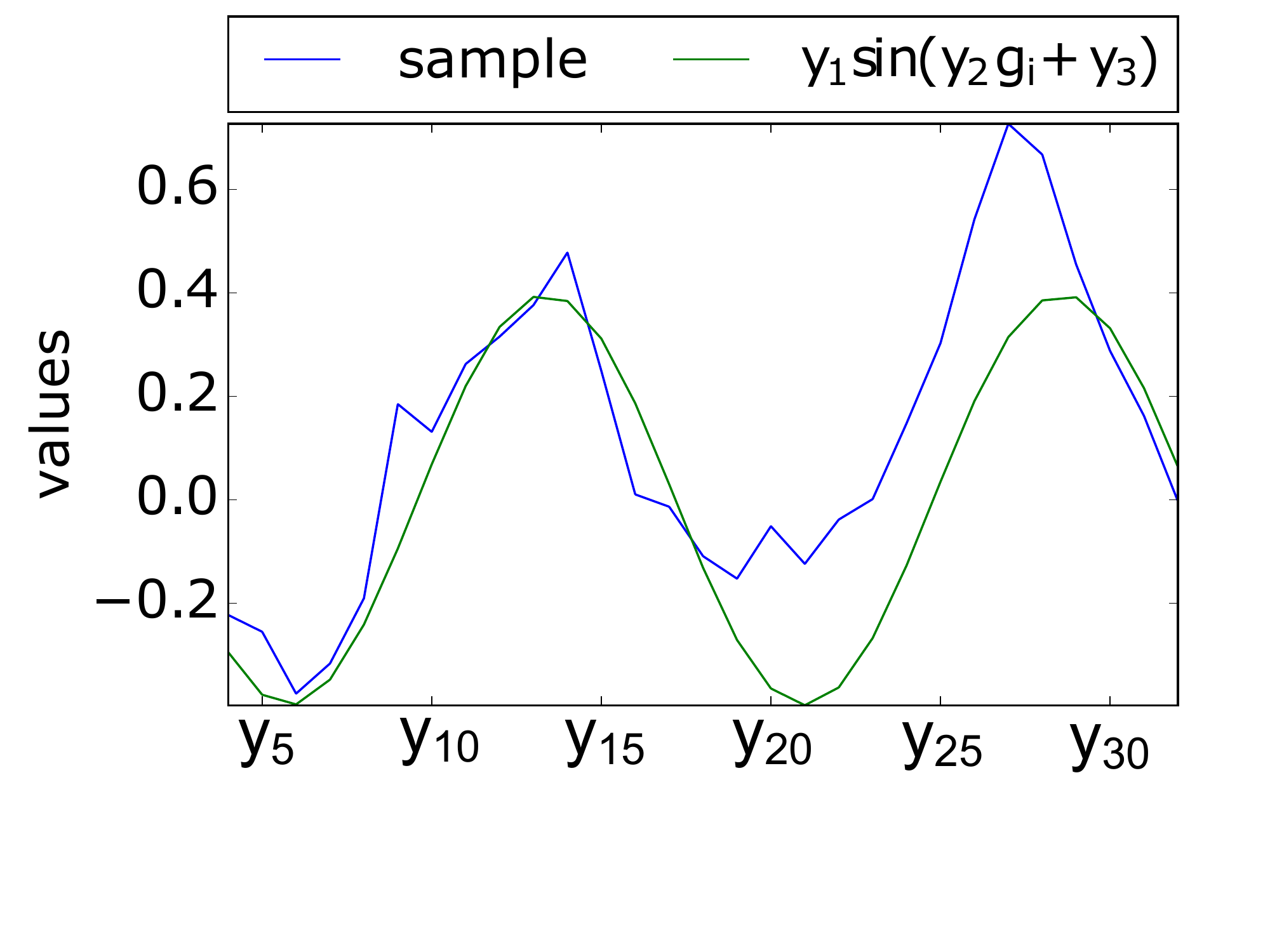}
    \end{subfigure}
    \begin{subfigure}[b]{0.24\textwidth}
        \includegraphics[width=\textwidth]{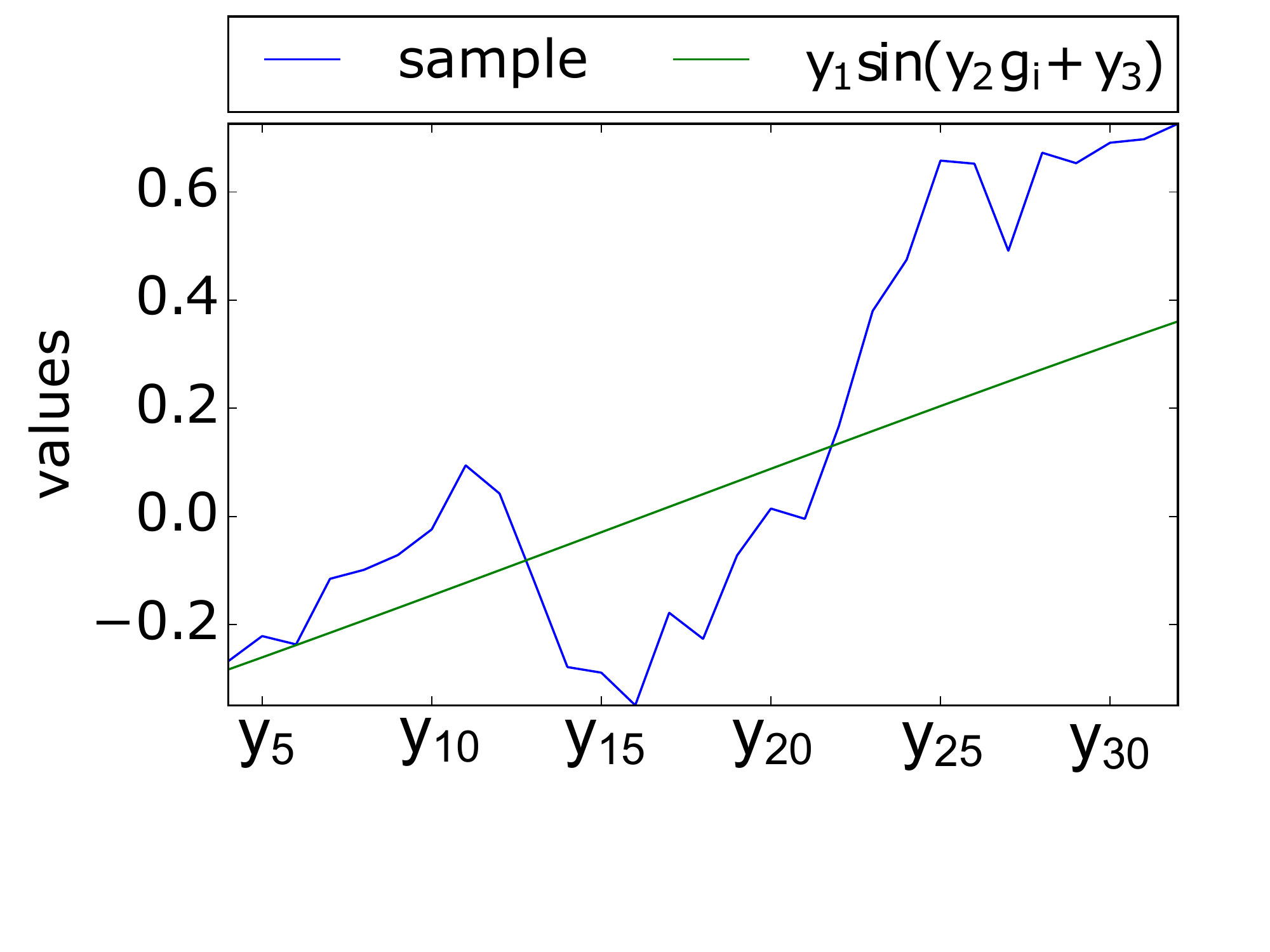}
    \end{subfigure}
    \begin{subfigure}[b]{0.24\textwidth}
        \includegraphics[width=\textwidth]{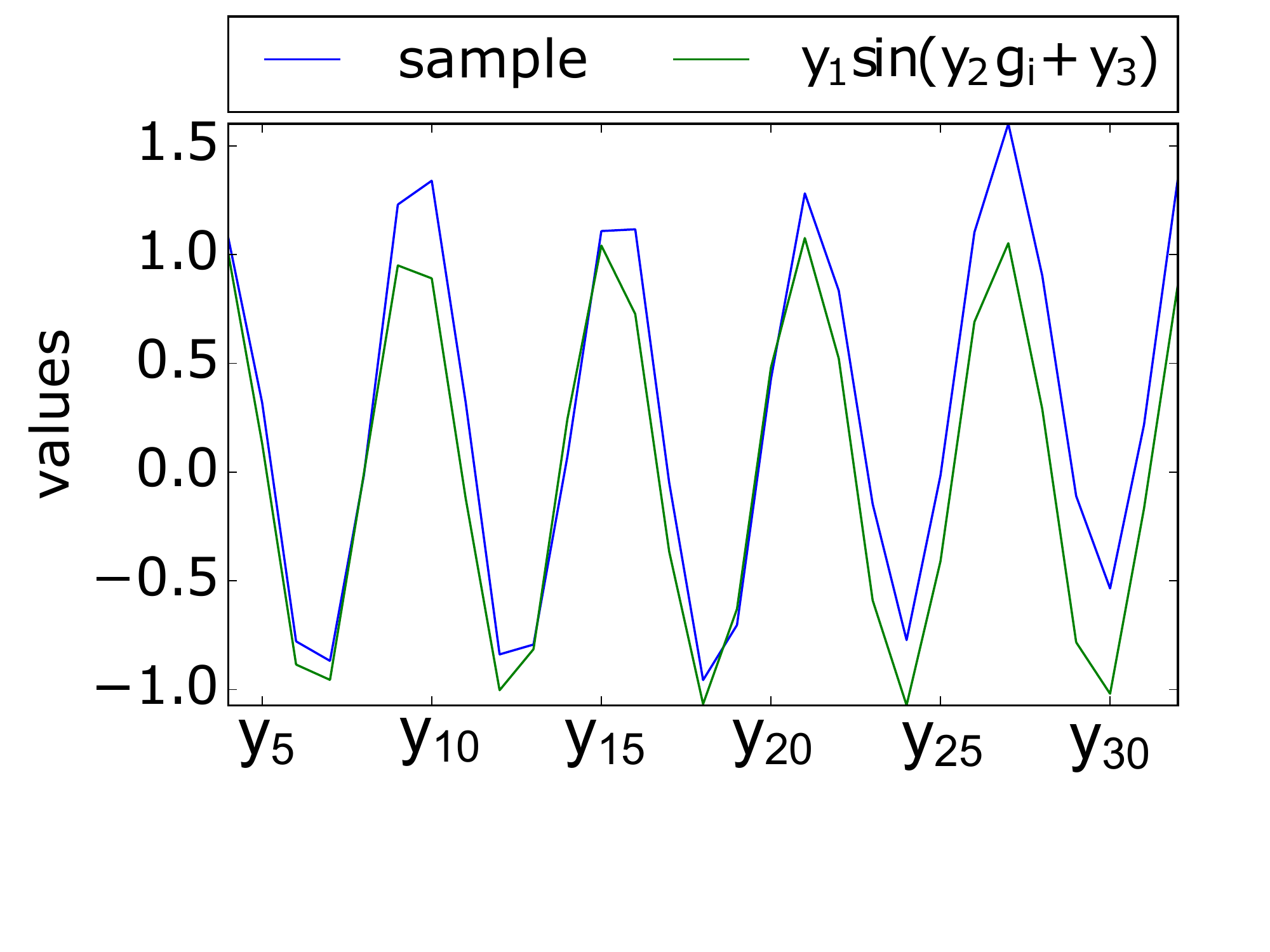}
    \end{subfigure}
    \vspace{-9mm}
    \caption{\texttt{RNN+4xSRNN+Re} \texttt{\&} \texttt{RAM} model samples. Each plot shows a single sample. We plot the sample values of unpermuted dimensions $y_4, \ldots, y_{32}\, |\, y_1, y_2, y_3$ in blue and the expected value of these dimensions (i.e. without the Markovian noise) in green. One may see that the model is able to correctly capture both the sinusoidal and random walk behavior of our data. }
    \label{fig:mark_samples}
    \vspace{-4mm}
\end{figure*}

\vspace{-3mm}
\paragraph{Methods}
We study the performance of various instantiation of TANs using different combinations of conditional models $p\left(q_i(x)\, |\, h_{i}\right)$ and various transformations $q(\cdot)$. 
%We study the performance of various combinations of conditional models and transformation. That is, we consider various models for the conditionals $p\left(q_i(x)\, |\, h_{i}\right)$ and various transformations $q(\cdot)$ \eqq{eq:nll}.
In particular the following conditional models were considered: \texttt{LAM}, \texttt{RAM}, \texttt{Tied}, \texttt{MultiInd}, and \texttt{SingleInd}. Here, \texttt{LAM}, \texttt{RAM}, and \texttt{Tied} are as described in equations \eqq{eq:untied}, \eqq{eq:rnn_model}, and \eqq{eq:nade}, respectively. \texttt{MultiInd} takes $p\left(q_{i}(x)\, |\, h_i\right)$ to be $p\left(q_i(x)\, |\, \MM(\theta_i)\right)$, that is we shall use $d$ distinct independent mixtures to model the transformed covariates. Similarly, \texttt{SingleInd} takes $p\left(q_i(x)\, |\, h_i\right)$ to be $p\left(q_i(x)\right)$, the density of a standard single component.
For transformations we considered: \texttt{None}, \texttt{RNN}, \texttt{2xRNN}, \texttt{4xAdd+Re}, \texttt{4xSRNN+Re}, \texttt{RNN+4xAdd+Re}, and \texttt{RNN+4xSRNN+Re}. \texttt{None} indicates that no transformation of variables was performed. \texttt{RNN} and \texttt{2xRNN} perform a single recurrent transformation \eqref{eq:rnn_trans}, and two recurrent transformations with a reversal permutation in between, respectively. Following \citep{dinh1}, \texttt{4xAdd+Re} performs four additive coupling transformations \eqq{eq:add_coup} with reversal permutations in between followed by a final element-wise rescaling: $x \mapsto x * \exp(s)$, where $s$ is a learned variable. Similarly, \texttt{4xSRNN+Re}, instead performs four recurrent shift transformations \eqref{eq:rnn_shift}. \texttt{RNN+4xAdd+Re}, and \texttt{RNN+4xSRNN+Re} are as before, but performing an initial recurrent transformation. Furthermore, we also considered performing an initial linear transformation \eqref{eq:linear_trans}. We flag this by prepending an \texttt{L} to the transformation; e.g. \texttt{L RNN} denotes a linear transformation followed by a recurrent transformation.
%For each task we compare TANs performance with corresponding state-of-the-art models.

\vspace{-3mm}
\paragraph{Implementation}
Models were implemented in Tensorflow \citep{tensorflow}\footnote{See \url{https://github.com/lupalab/tan}.}. Both RAM conditional models as well as the RNN shift transformation make use of the standard \texttt{GRUCell} GRU implementation. We take the mixture models of conditionals \eqq{eq:auto_MM} to be mixtures of 40 Gaussians. We optimize all models using the \texttt{AdamOptimizer} \citep{kingma2014adam} with an initial learning rate of $0.005$. Training consisted of $30\,000$ iterations, with mini-batches of size $256$. The learning rate was decreased by a factor of $0.1$, or $0.5$ (chosen via a validation set) every $5\,000$ iterations. Gradient clipping with a norm of $1$ was used. After training, the best iteration according to the validation set loss was used to produce the test set results.

\vspace{-2mm}
\subsection{Synthetic\label{sec:exptSynth}}
\vspace{-2mm}
%We perform a thorough empirical analysis over synthetic data.
To showcase the strengths of TANs and short-comings of only conditional models \& only transformations, we carefully construct two synthetic datasets% as discussed below.

\vspace{-3mm}
\paragraph{Data Generation}
%\subsubsection{Markovian Data}
Our first dataset, consisting of a %n easy-to-visualize 
Markovian structure that features several exploitable correlations among covariates, is constructed as: $y_1, y_2, y_3 \sim \mathcal{N}(0,1)$ and $y_i\,|\,y_{i-1},\ldots,y_1 \sim f(i, y_1, y_2, y_3) + \epsilon_i$ for $i>3$ where $\epsilon_i \sim \mathcal{N}(\epsilon_{i-1},\sigma)$, $f(i, y_1, y_2, x_3) = y_1 \sin( y_2 g_i + y_3 )$, and $g_i$'s are equi-spaced points on the unit interval.
That is, instances are sampled using random draws of amplitude, frequency, and shift covariates $y_1, y_2, y_3$, which determine the mean of the other covariates, $y_1 \sin( y_2 g_i + y_3 )$, stemming from function evaluations on a grid, and random noise $\epsilon_i$ with a Gaussian random walk. 
The resulting instances contain many correlations as visualized in \reffig{fig:mark_samples}.
%are easy to visualize (\reffig{fig:mark_samples}), and 
%contain many correlations among covariates 
%(for instance, $y_4$ is highly informative of $y_5$).
To further exemplify the importance of employing conditional and transformations in tandem, we construct a second dataset with much fewer correlations. In particular, we use a star-structured graphical model where fringe nodes are very uninformative of each-other and estimating the distribution of the fringe vertices are difficult without conditioning on all the center nodes. To construct the dataset: 
divide the covariates into disjoint center and vertex sets $C = \{1, \ldots, 4\},\, V = \{5, \ldots, d\}$ respectively. For center nodes $j \in C$, $y_j \sim \mathcal{N}(0,1)$. Then, for $j \in V$, $y_j \sim \mathcal{N}(f_j(w_j^{T}y_C),\sigma)$ where $f_j$ is a fixed step function with 32 intervals, $w_j\in\R^4$ is a fixed vector, and $y_C= (y_1, y_2, y_3, y_4)$. In both datasets, to test robustness to correlations from distant (by index) covariates, we observe covariates that are shuffled using a fixed permutation $\pi$ chosen ahead of time: $x = (y_{\pi_1}, \ldots, y_{\pi_d})$. We take $d=32$, and the number of training instances to be $100\,000$.
%We note that this dataset poses a difficult density estimation problem since the distribution of each of the fringe vertices will be considerably different from each other, the fringe vertices are largely uninformative from one another, and the distribution of the fringe vertices are difficult to estimate without conditioning on all the center nodes.

\vspace{-3mm}
\paragraph{Observations}
We detail the mean log-likelihoods on a test set for TANs using various combinations of conditional models and transformations in Appendix, \reftab{tab:markov} and \reftab{tab:star} respectively. 
%Note that the performance of previous one-prong approaches that considered a complex conditional model with simple or no transformation and vice-versa are illustrated by \NADE (NADE), \NICE (NICE) models, as well as by the entire row corresponding to \None\ transformation and the \MultiInd and \SingleInd columns. 
We see that both \LAM and \RAM conditionals are providing most of the top models. We observe good samples from the best performing model as shown in \reffig{fig:mark_samples}. 
%Here we also observe relatively good performance stemming from \MultiInd conditionals with more complex transformations.
Particularly in second dataset, simpler conditional methods are unable to model the data well, suggesting that the complicated dependencies need a two-prong TAN approach. We observe a similar pattern when learning over the star data with $d=128$ (see Appendix, \reftab{tab:star128}).

\begin{table*}[t]
\small
\vspace{-2mm}
\newcommand{\ra}[1]{\renewcommand{\arraystretch}{#1}}
\ra{1.1}
    \centering
    \caption{Average test log-likelihood comparison of TANs with baselines MADE, Real NVP, MAF as reported by \cite{papamakarios2017masked}. For TANs the best model is picked using validation dataset and are reported here. Parenthesized numbers indicate number of transformations used. Standard errors with $2\sigma$ are shown. Largest values per dataset are shown in \textbf{bold}. \label{tab:maf_llks}}
    \vspace{1mm}
    %\begin{tabular}{@{} L{20mm} R{20mm} R{20mm} R{20mm} R{20mm} R{20mm} @{}} 
    \begin{tabular}{@{} l c c c c c @{}} 
    \toprule
    & \makecell{\textsf{POWER}\\ {\tiny \textsf{d=6; N=2,049,280}}}
    & \makecell{\hphantom{33}\textsf{GAS}\\ {\tiny \hphantom{33}\textsf{d=8; N=1,052,065}}}
    & \makecell{\hphantom{33}\textsf{HEPMASS}\\ {\tiny \hphantom{333}\textsf{d=21; N=525,123}}}
    & \makecell{\hphantom{3}\textsf{MINIBOONE}\\ {\tiny \hphantom{333}\textsf{d=43; N=36,488}}}
    & \makecell{\hphantom{3}\textsf{BSDS300}\\ {\tiny \hphantom{33}\textsf{d=63; N=1,300,000}}}
    \\ 

    \midrule
%        Gaussian & -7.74 $\pm$ 0.02  &-3.58 $\pm$ 0.75 & -27.93 $\pm$ 0.02 & -37.24 $\pm$ 1.07 & 96.67 $\pm$ 0.25 \\
%        & & & & & \\
        MADE &-3.08 $\pm$ 0.03 & 3.56 $\pm$ 0.04 &  -20.98 $\pm$ 0.02 &  -15.59 $\pm$ 0.50 &  148.85 $\pm$ 0.28 \\
MADE MoG &  0.40 $\pm$ 0.01 &  8.47 $\pm$ 0.02 &  -15.15 $\pm$ 0.02 &  -12.27 $\pm$ 0.47 &  153.71 $\pm$ 0.28 \\
% & & & & & \\
Real NVP (5) &  -0.02 $\pm$ 0.01 &  4.78 $\pm$ 1.80 &  -19.62 $\pm$ 0.02 &  -13.55 $\pm$ 0.49 &  152.97 $\pm$ 0.28 \\
Real NVP (10) &  0.17 $\pm$ 0.01 &  8.33 $\pm$ 0.14 &  -18.71 $\pm$ 0.02 &  -13.84 $\pm$ 0.52 &  153.28 $\pm$ 1.78 \\
% & & & & & \\
MAF (5) &  0.14 $\pm$ 0.01 &  9.07 $\pm$ 0.02 &  -17.70 $\pm$ 0.02 &  -11.75 $\pm$ 0.44 &  155.69 $\pm$ 0.28 \\
MAF (10) &  0.24 $\pm$ 0.01 &  10.08 $\pm$ 0.02 &  -17.73 $\pm$ 0.02 &  -12.24 $\pm$ 0.45 &  154.93 $\pm$ 0.28 \\
MAF MoG (5) &  0.30 $\pm$ 0.01 &  9.59 $\pm$ 0.02 &  -17.39 $\pm$ 0.02 &  -11.68 $\pm$ 0.44 &  156.36 $\pm$ 0.28 \\
\midrule
%\vspace{-3mm} & & & & & \\
%TAN & $\bm{0.48 \pm 0.01}$ & $\bm{11.19 \pm 0.02}$ & $\bm{-15.12 \pm 0.02}$ & $\bm{-11.01 \pm 0.48}$ & $\bm{157.03 \pm 0.07}$ \\
% & {\tiny\makecell{L RNN+4xAdd+Re\\ \& RAM}} & {\tiny\makecell{L RNN+4xSRNN+Re\\ \& RAM}} & {\tiny\makecell{L RNN\\ \& RAM}} & {\tiny\makecell{4xSRNN+Re\\ \& RAM}} & {\tiny\makecell{L RNN+4xSRNN+Re\\ \& RAM}} \\
TAN & $\bm{0.60 \pm 0.01}$ & $\bm{12.06 \pm 0.02}$ & $\bm{-13.78 \pm 0.02}$ & $\bm{-11.01 \pm 0.48}$ & $\bm{159.80 \pm 0.07}$ \\
 & {\tiny\makecell{5x L+ReLU+SRNN+Re\\ \& RAM}} & {\tiny\makecell{5x L+ReLU+SRNN+Re\\ \& RAM}} & {\tiny\makecell{5x L+ReLU+SRNN+Re\\ \& RAM}} & {\tiny\makecell{4xSRNN + Re\\ \& RAM}} & {\tiny\makecell{5x L+ReLU+SRNN+Re\\ \& RAM}} \\
 \bottomrule
\end{tabular}
\vspace{-2mm}
\end{table*}
\normalsize

\vspace{-2mm}
\subsection{Efficacy on Real World Data\label{sec:exptReal}}
\vspace{-2mm}
We performed several real-world data experiments and compared to several state-of-the-art density estimation methods to substantially improved performance of TAN. 

\vspace{-3mm}
\paragraph{Datasets} 
We carefully followed \cite{papamakarios2017masked} and code \citep{mafcode} to ensure that we operated over the same instances and covariates for each of the datasets considered in \cite{papamakarios2017masked}. Specifically we performed unconditional density estimation on four datasets from UCI machine learning repository\footnote{\url{http://archive.ics.uci.edu/ml/}}: \texttt{power}: Containing electric power consumption in a household over 47 months. \texttt{gas}: Readings of 16 chemical sensors exposed to gas mixtures. \texttt{hepmass}: Describing Monte Carlo simulations for high energy physics experiments. \texttt{minibone}: Containing examples of electron neutrino and muon neutrino. We also used \texttt{BSDS300} which were obtained from extracting random $8 \times 8$ monochrome patches from the BSDS300 datasets of natural images \citep{martin2001database}.
These are multivariate datasets from a varied set of sources meant to provide a broad picture of performance across different domains. 
Here, we used a batch size of 1024 with 60K training iterations. We saw great performance by using multiple successions of a linear transformation, followed by an element-wise leaky transformation (as in eq. \ref{eq:rnn_trans}), a recurrent shift transformation \eqref{eq:rnn_shift}, and an element-wise rescale transformation. Thus in addition, we used a model with 5 such stacked transformations (\texttt{\small 5x L+ReLU+SRNN+Re}).
Further, to demonstrate that our proposed models can even be used to model high dimensional data and produce coherent samples, we consider image modeling task, treating each image as a flattened vector. We consider $28 \times 28$ grayscale images of MNIST digits and $32 \times 32$ natural colored images of CIFAR-10. Following \citet{dinh1}, we dequantize pixel values by adding noise and rescaling.% to the unit interval.

\vspace{-3mm}
\paragraph{Metric} 
\begin{wrapfigure}{r}{0.36\columnwidth}
    \vspace{-4mm}
    \centering
    \includegraphics[trim={0cm 3mm 0cm 0},clip, width=\linewidth]{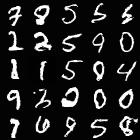}
    \vspace{-6.5mm}
    \caption{Samples from best TAN model.}
    \label{fig:mnist_samples}
    \vspace{-3mm}
\end{wrapfigure}
We use the average test log-likelihoods of the best TAN model selected using a validation set and compare to values reported by \cite{papamakarios2017masked} for MADE \citep{germain2015made}, Real NVP \citep{dinh2}, and MAF \citep{papamakarios2017masked} methods for each dataset.
%Also, we compared to MAF MoG, which uses a MAF transformation of variables \citep{papamakarios2017masked} with a MADE MoG conditional model \citep{germain2015made}.
For images, we use transformed version of test log-likelihood, called bits per pixel, which is more popular. 
In order to calculate bits per pixel, we need to convert the densities returned by a model back to image space in the range [0, 255], for which we use the same logit mapping provided
in \citet[Appendix E.2]{papamakarios2017masked}. 
%This transformation will lessen boundary effects and keep pixel values inside a valid range.

\begin{figure}[t]
   \vspace{-1mm}
    \centering
    \begin{subfigure}[b]{0.56\columnwidth}
        \includegraphics[trim={0.4cm 0 0.4cm 0},clip, height=36mm]{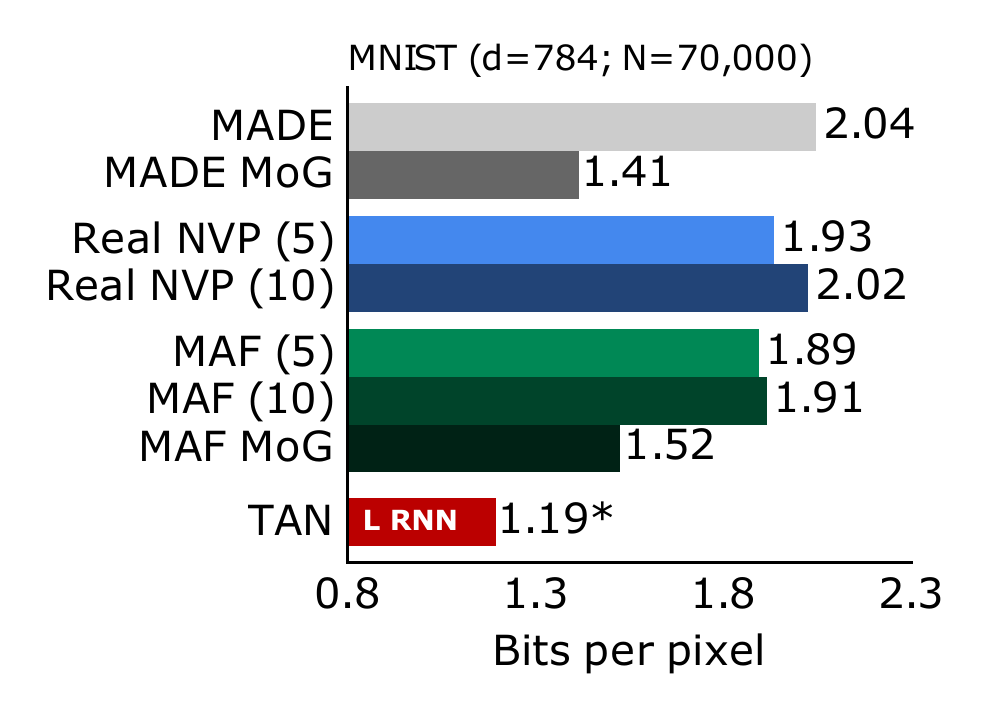}
    \end{subfigure}
    \begin{subfigure}[b]{0.40\columnwidth}
        \includegraphics[trim={0.5cm 0 0.0cm 0},clip,height=36mm]{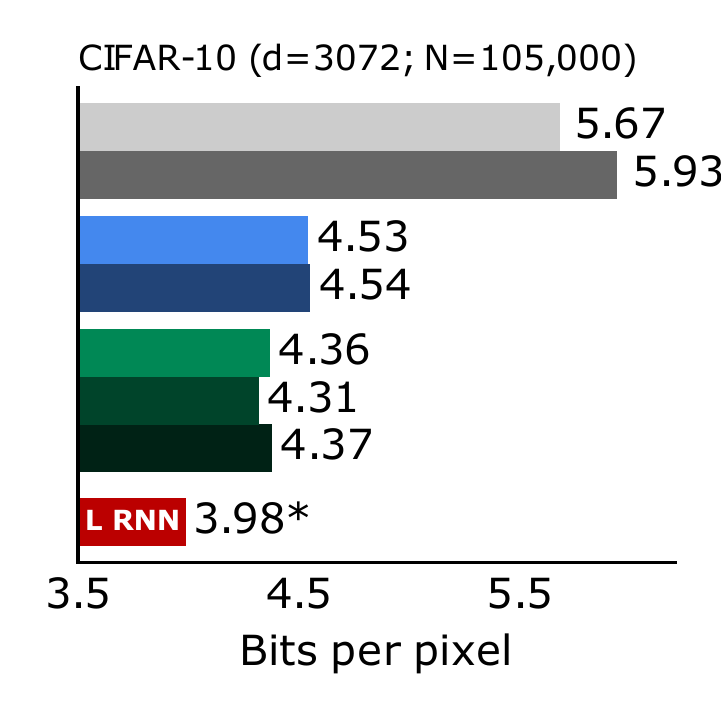}
    \end{subfigure}
    \vspace{-5mm}
    \caption{Bits per pixel for models (lower is better) using logit transforms on MNIST $\&$ CIFAR-10. MADE, Real NVP, and MAF values are as reported by \cite{papamakarios2017masked}. % Standard errors with $2\sigma$ are shown. 
    The best achieved value is denoted by *. }
    \label{fig:comp}
    \vspace{-5mm}
\end{figure}

\vspace{-3mm}
\paragraph{Observations}
\reftab{tab:maf_llks} and \reffig{fig:comp} shows our results on various multivariate datasets and images respectively, with error bars computed over 5 runs. 
As can be seen, our TAN models are considerably outperforming other state-of-the-art
methods across all multivariate as well as image datasets, justifying our claim of utilizing both complex transformations and conditionals.
Furthermore, we plot samples for MNIST case in ~\reffig{fig:mnist_samples}. We see that TAN is able to capture the structure of digits with very few artifacts in samples, which is also reflected in the likelihoods.

% \footnote{The \texttt{HEPMASS} dataset uses a subset of instances and covariates from \texttt{hepmass}.}

\begin{figure*}[t]
\vspace{-2mm}
    \centering
    %%%%%%%%%%%%%%%%%%%%%%%%%%%%%%%%%%%%%%%%%%%%%%%%%%%%%%%%%%%%%
    \begin{subfigure}[b]{0.29\textwidth}
        \includegraphics[trim={0.4cm 0 0.0cm 0},clip, height=42mm]{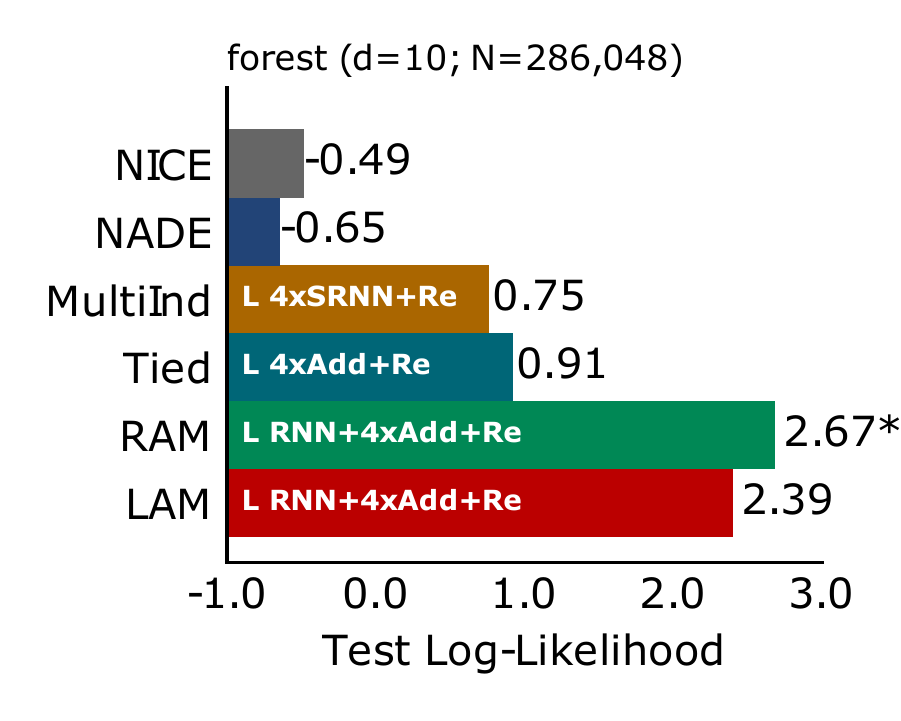}
    \end{subfigure}\hfill
    \begin{subfigure}[b]{0.225\textwidth}
            \includegraphics[trim={0.4cm 0 0.1cm 0},clip, height=42mm]{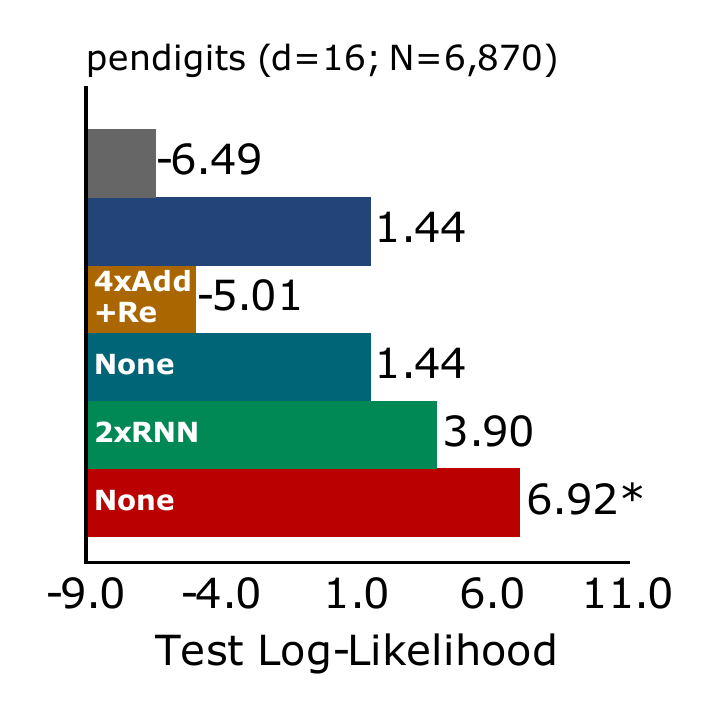}
    \end{subfigure}\hfill
    \begin{subfigure}[b]{0.225\textwidth}
            \includegraphics[trim={0.4cm 0 0.4cm 0},clip, height=42mm]{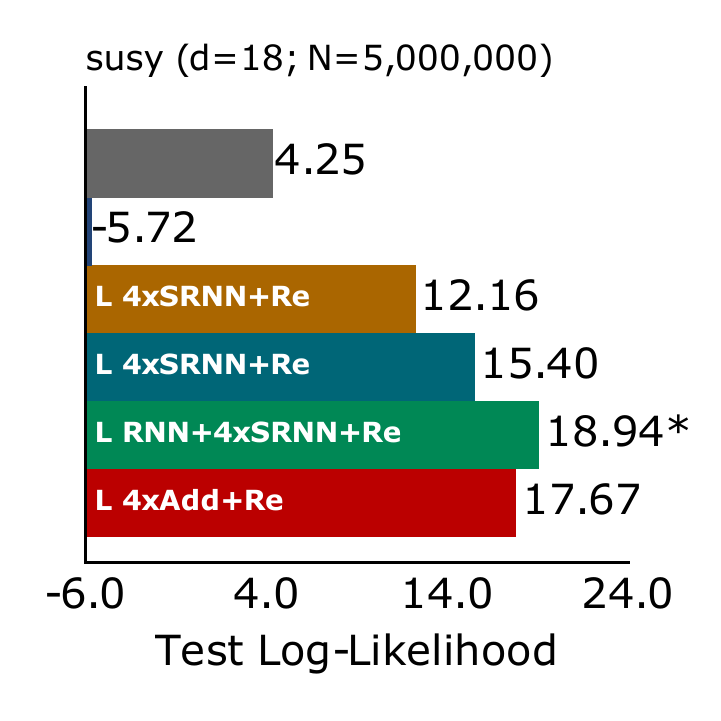}
    \end{subfigure}\hfill
    \begin{subfigure}[b]{0.225\textwidth}
        \includegraphics[trim={0.4cm 0 0.4cm 0},clip, height=42mm]{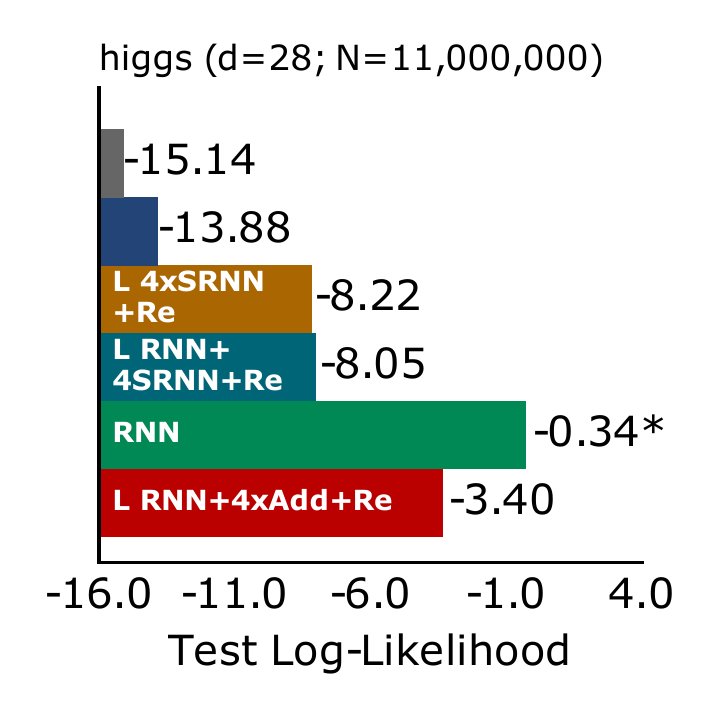}
    \end{subfigure}\\
    \vspace{-2mm}
	%%%%%%%%%%%%%%%%%%%%%%%%%%%%%%%%%%%%%%%%%%%%%%%%%%%%%%%%%%%%%
  \begin{subfigure}[b]{0.29\textwidth}
        \includegraphics[trim={0.4cm 0 0.0cm 0},clip, height=42mm]{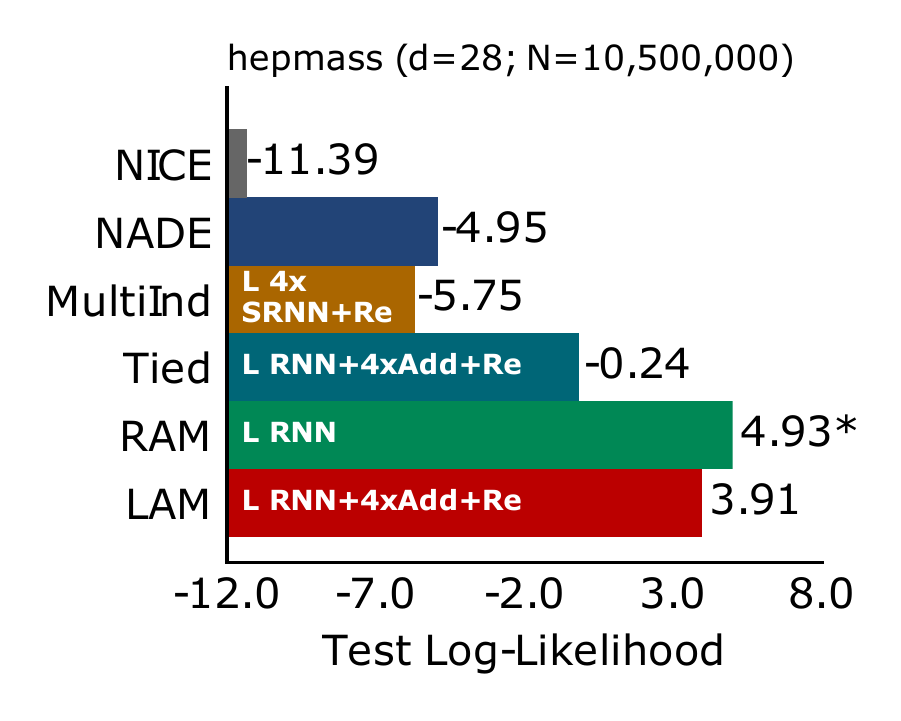}
    \end{subfigure}
    \begin{subfigure}[b]{0.225\textwidth}
            \includegraphics[trim={0.4cm 0 0.1cm 0},clip, height=42mm]{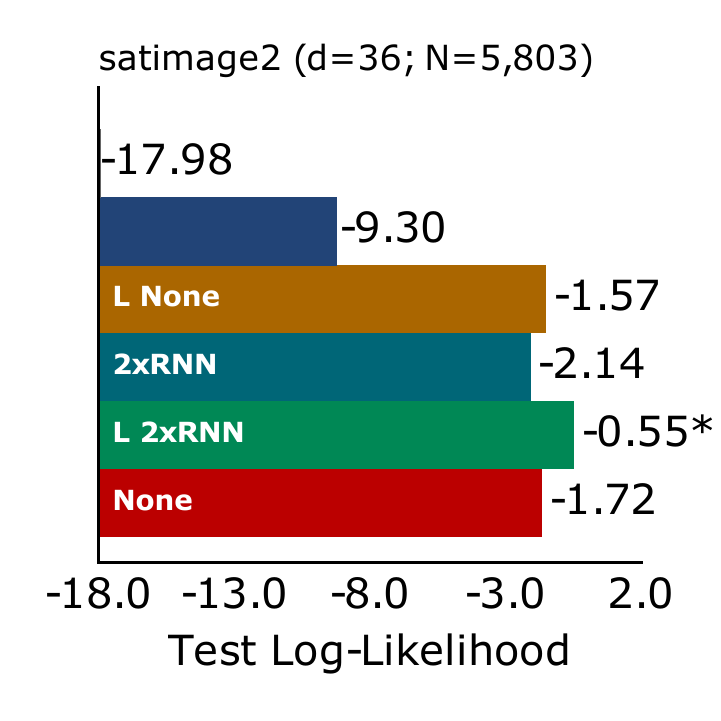}
    \end{subfigure}
    \begin{subfigure}[b]{0.225\textwidth}
            \includegraphics[trim={0.4cm 0 0.4cm 0},clip, height=42mm]{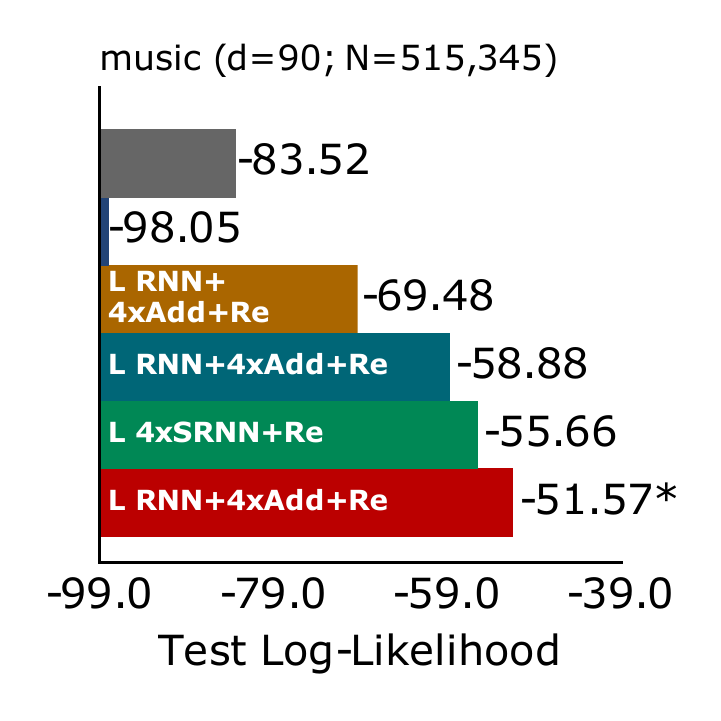}
    \end{subfigure}
    \begin{subfigure}[b]{0.225\textwidth}
        \includegraphics[trim={0.4cm 0 0.4cm 0},clip, height=42mm]{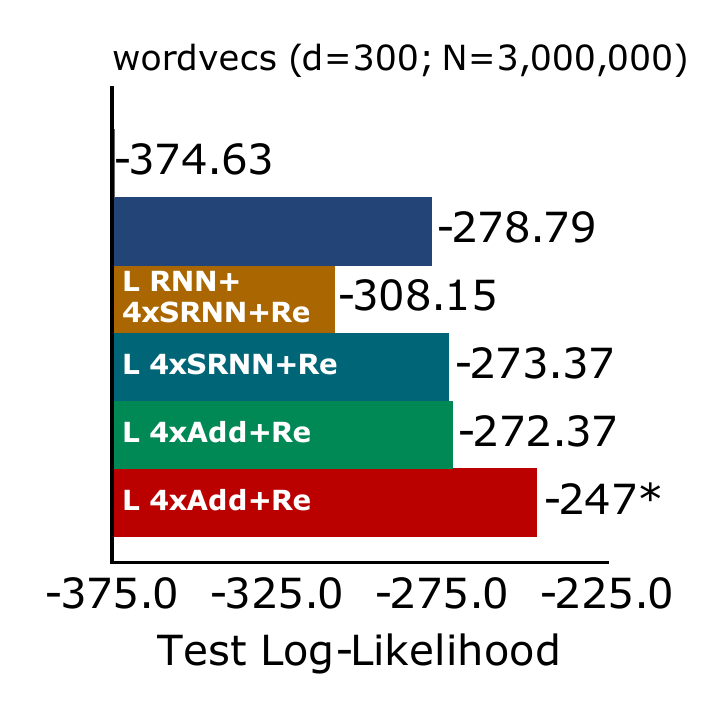}
    \end{subfigure}
    %%%%%%%%%%%%%%%%%%%%%%%%%%%%%%%%%%%%%%%%%%%%%%%%%%%%%%%%%%%%%
    \vspace{-5mm}
    \caption{Ablation Study of various components TAN. For each dataset and each conditional model, top transformations is selected using log-likelihoods on a validation set. The picked transformation is reported within the bars for each conditional. $*$ denotes the best model for each dataset picked by validation. Simple conditional \texttt{MultiInd}, always lags behind sophisticated conditionals such as \LAM \& \RAM. \label{fig:llks}}
    \label{fig:real-dataset-plots}
    \vspace{-3mm}
\end{figure*}

\vspace{-2mm}
\subsection{Ablation Study\label{sec:exptAbl}}
\vspace{-2mm}
To study how different components of the models affect the log-likelihood, we perform a comprehensive ablation study across different datasets.

\vspace{-3mm}
\paragraph{Datasets} 
We used multiple datasets from the UCI machine learning repository\footnote{\url{http://archive.ics.uci.edu/ml/}} and Stony Brook outlier detection datasets collection (ODDS)\footnote{\url{http://odds.cs.stonybrook.edu}}
to evaluate log-likelihoods on test data. 
%These are multivariate datasets from varied set of sources meant to provide a complete picture of performance general data modeling tasks. 
Broadly, the datasets can be divided into:
\textbf{Particle acceleration}: \texttt{higgs}, \texttt{hepmass}, and \texttt{susy} datasets where generated for high-energy physics experiments using Monte Carlo simulations;
\textbf{Music}: The \texttt{music} dataset contains timbre features from the million song dataset of mostly commercial western song tracks from the year 1922 to 2011; \citep{millisongs}.
\textbf{Word2Vec}: \texttt{wordvecs} consists of 3 million words from a Google News corpus. Each word represented as a 300 dimensional vector trained using a word2vec model\footnote{\url{https://code.google.com/archive/p/word2vec/}}.
\textbf{ODDS datasets}: We used several ODDS datasets--\texttt{forest}, \texttt{pendigits}, \texttt{satimage2}. These are multivariate datasets from varied set of sources meant to provide a broad picture of performance across anomaly detection tasks. 
To not penalize models for low likelihoods on outliers in ODDS, we removed anomalies from test sets. % when reporting log-likelihoods.

As noted in \citep{dinh1}, data degeneracies and other corner-cases may lead to arbitrarily low negative log-likelihoods. 
%To avoid such complications, we remove discrete features, standardize, and add independent Gaussian noise with a standard deviation of $0.01$ to training sets. 
Thus, we remove discrete features, standardize, and add Gaussian noise (stddev of $0.01$) to training sets. 

\vspace{-3mm}
\paragraph{Observations}
We report average test log-likelihoods in \reffig{fig:real-dataset-plots} for each dataset and conditional model for the top transformations picked on a validation dataset.
The tables with test log-likelihoods for all combinations of conditional models and transformations for each dataset is in Appendix \reftab{tab:forestllk}-\ref{tab:music}.
%Methods that only consider complex transformations or condition models are illustrated in the entire row corresponding to the \None $\ $transformation and the \MultiInd and \SingleInd columns, respectively. 
We observe that the best performing models in real-world datasets
are those that incorporate a flexible transformation \emph{and} conditional model. In fact, the best model in each of the datasets considered
%(according to validation dataset)
always has \LAM or \RAM autoregressive components. 
%Hence, validation across models would always select one of these methods. 
Each row of these tables show that using a complex conditional is always better than using restricted, independent conditionals. Similarly, each column of the table shows that for a given conditional, it is better to pick a complex transformation rather than having no transformation. % at all.
It is interesting to note that many of these top models also contain a linear transformation. Of course, linear transformations of variables are common to most parametric models, however they have been under-explored in the context of autoregressive density estimation. Our methodology for efficiently learning linear transformations coupled with their strong empirical performance encourages their inclusion in autoregressive models for most datasets.

\begin{figure*}[t]
    \centering
    \vspace{-3mm}
    \begin{turn}{90}\bf\qquad\;\;\; Original\end{turn}
    \includegraphics[width=0.15\textwidth]{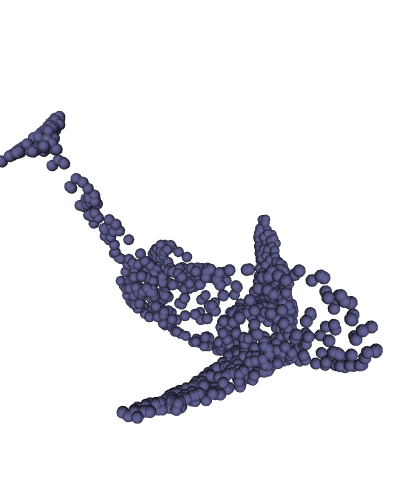}\hfill
    \includegraphics[width=0.15\textwidth]{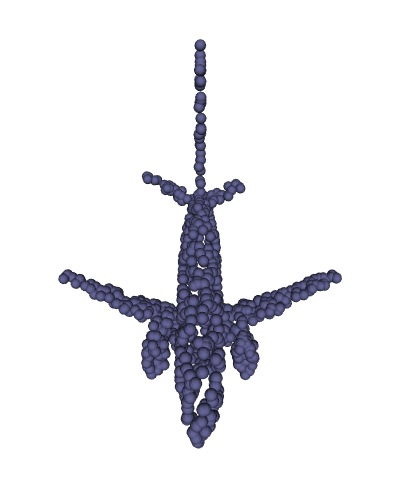}\hfill
    \includegraphics[width=0.15\textwidth]{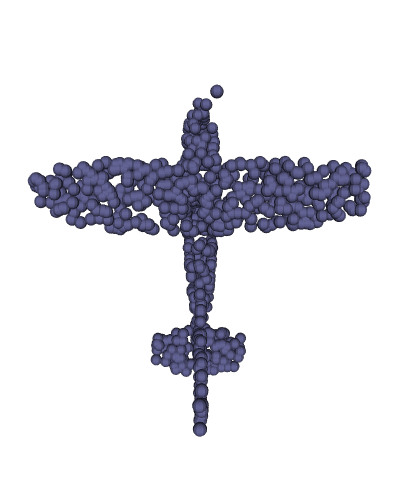}\hfill
    \includegraphics[trim={-25mm -25mm -25mm -25mm}, clip, width=0.15\textwidth]{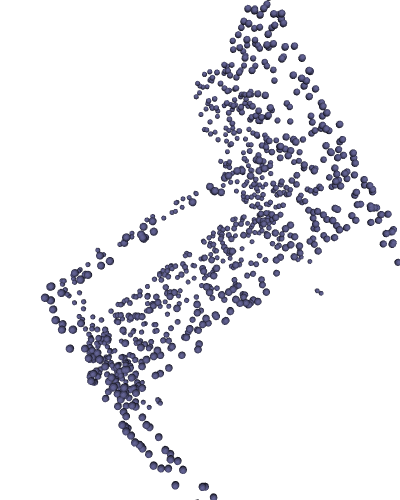}\hfill
    \includegraphics[trim={-20mm -20mm -20mm -20mm}, clip, width=0.15\textwidth]{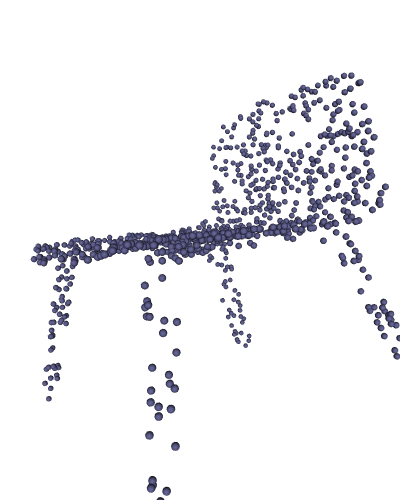}\hfill
    \includegraphics[trim={-15mm -15mm -15mm -15mm}, clip, width=0.15\textwidth]{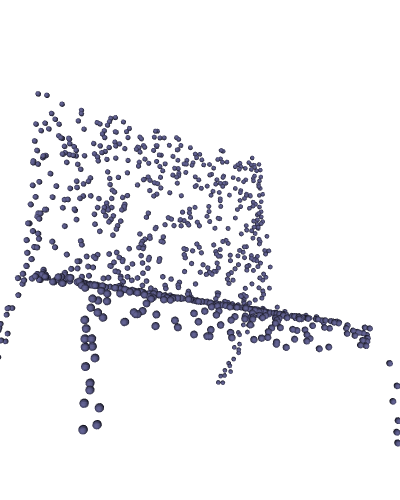}
    \vspace{-2mm}
    \\
    \begin{turn}{90}\bf\qquad\quad Samples\end{turn}
    \includegraphics[width=0.15\textwidth]{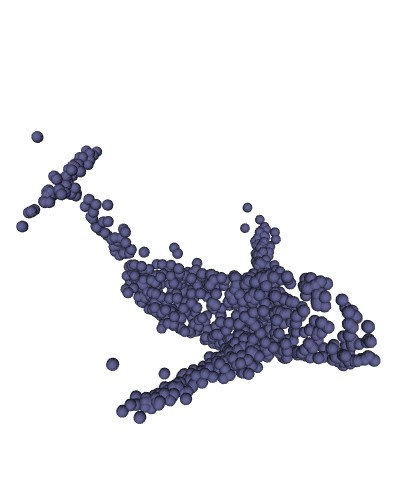}\hfill
    \includegraphics[width=0.15\textwidth]{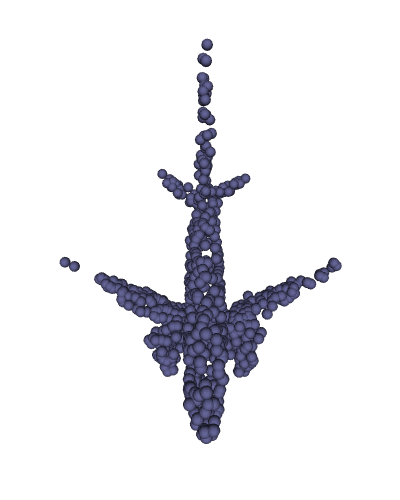}\hfill
    \includegraphics[width=0.15\textwidth]{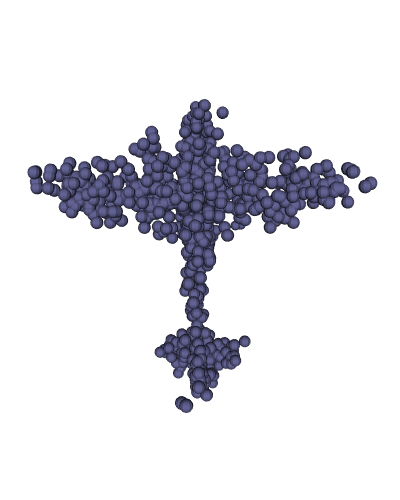}\hfill
    \includegraphics[trim={-25mm -35mm -25mm -25mm}, clip, width=0.15\textwidth]{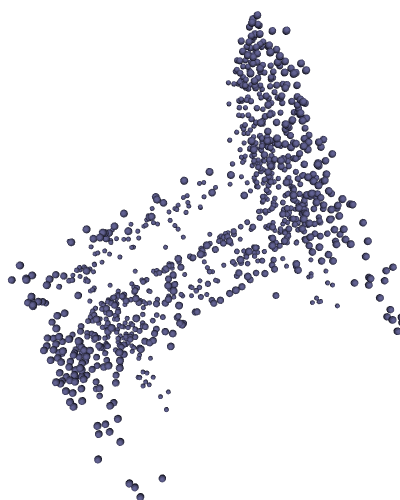}\hfill
    \includegraphics[trim={-20mm -35mm -20mm -20mm}, clip, width=0.15\textwidth]{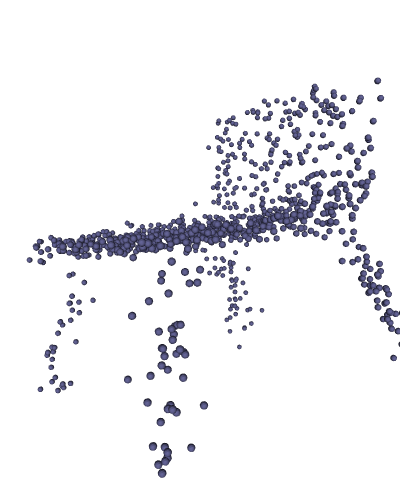}\hfill
    \includegraphics[trim={-15mm -25mm -15mm -15mm}, clip, width=0.15\textwidth]{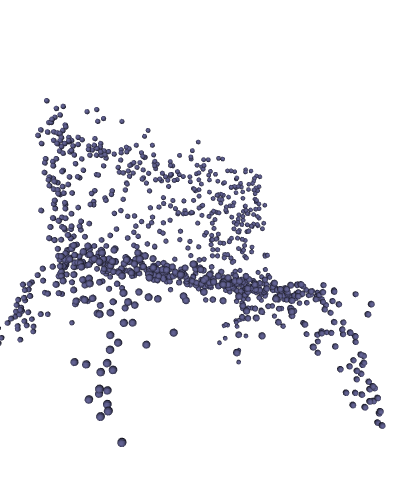}
    \vspace{-6mm}
    \caption{Qualitative samples obtained from TANs for the task of learning parametric family of distributions where we treat each category of objects as a family and each point cloud for an object as the sample set. Top row shows unseen test point clouds and bottom row represents samples produced from TANs for these inputs.
Presence of few artifacts in samples of unseen objects indicates a good fit.}
    \label{fig:pc_samples}
    \vspace{-3mm}
\end{figure*}

%Finally, we want to pick the ``overall'' winning combination of transformations and conditionals. 
Finally, we pick the ``overall'' winning combination of transformations and conditionals. 
For this we compute the fraction of the top likelihood achieved by each transformation $t$ and conditional model $m$ for dataset $D$: $s(t, m, D) = \exp(l_{t, m, D})/\max_{a, b} \exp(l_{a, b, D})$, where $l_{t, m, D}$ is the test log-likelihood for $t, m$ on $D$. 
%That is, $s(t, m, D)$ is the percentage of the top likelihood achieved by transformation $t$ and conditional model $m$ on dataset $D$. 
We then average $S$ over the datasets: $S(t,m) = \frac{1}{T}\sum_{D} S(t, m, D)$, where $T$ is the total number of datasets and reported all these score in Appendix ~\reftab{tab:combined_scored}. This provides a summary of which models performed better over multiple datasets. In other words, the closer this score is to 1 for a model means the more datasets for which the model is the best performer. We see that \RAM conditional with \texttt{L RNN} transformation, and \LAM conditional with \texttt{L RNN+4xAdd+Re} were the two best performers.
%with high scores of <> and <> respectively. 

\begin{figure}[!b]
\vspace{-6mm}
\centering
\includegraphics[width=0.6\columnwidth]{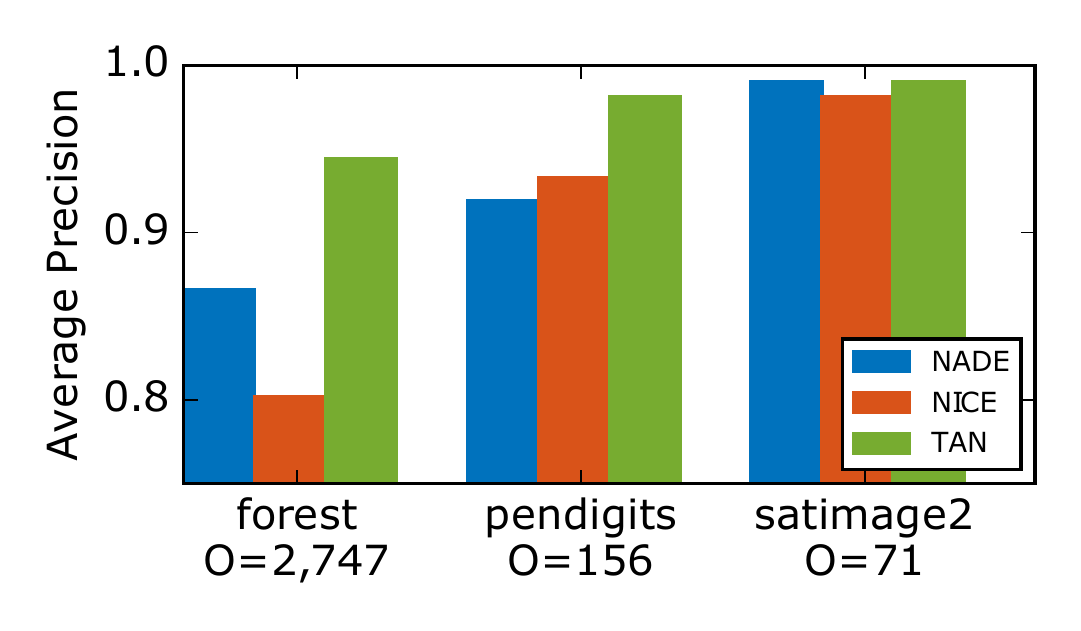}
\vspace{-6mm}
\caption{Average precision score on outlier detection datasets. For each dataset, the best performing TAN model picked using likelihood on a validation set, is shown. \label{fig:ocd}}
\vspace{-3mm}
\end{figure}

\vspace{-2mm}
\subsection{Anomaly Detection\label{sec:exptAnom}}
\vspace{-2mm}
Next, we apply density estimates to anomaly detection. Typically anomalies or outliers are data-points that are unlikely given a dataset. In terms of density estimations, such a task is framed by identifying which instances in a dataset have a low corresponding density. That is, we shall label an instance $x$, as an anomaly if $\hat{p}(x) \leq t$, where $t\geq 0$ is some threshold and $\hat{p}$ is the density estimate based on training data. Note that this approach is trained in an unsupervised fashion. Density estimates were evaluated on test data with anomaly/non-anomaly labels on instances. We used thresholded log-likelihoods on the test set to compute precision and recall. 
We use the average-precision metric and show our results in \reffig{fig:ocd}. TAN performs the best on all three datasets.
Beyond providing another interesting use for our density estimates, seeing good performance in these outlier detection tasks further demonstrates that our models are learning semantically meaningful patterns.

\vspace{-2mm}
\subsection{Learning Parametric Family of Distributions\label{sec:exptEmbd}}
\vspace{-2mm}
To further demonstrate flexibility of TANs, we consider a new task of learning parametric family of distributions together. Suppose we have a family of density $\mathcal{P}_\theta$. 
We assume in training data there are $N$ sets $X_1, ...,X_N$ , where the $n$-th set $X_n = \{ x_{n,1}, ..., x_{n,m_n} \}$ consists of $m_n$ i.i.d. samples from
density $\mathcal{P}_{\theta_n}$, \ie $X_n$ is a set of sample points, and
$x_{n,j} \sim \mathcal{P}_{\theta_n}, j = 1, ..., m_n$. We assume that we do not have access to underlying true parameters $\theta_n$. 
% Moreover, not all parameters $\theta_n$ may be unique. 
We want to jointly learn the density estimate and parameterization of the sets to  predict even for sets coming from unseen values of $\theta$.

We achieve this with a novel approach that models each set $X_i$ with $p(\cdot | \phi(X_i))$ where $p$ is a shared TAN model for the family of distributions and $\phi(X_i)$ are a learned embedding (parameters) for the $i$th set with DeepSets \citep{zaheer2017deep}.
%We propose a solution to the problem by combining TAN with DeepSets \citep{zaheer2017deep}. 
In particular, we use a permutation invariant network of DeepSets parameterized by $W_1$ to extract the embedding $\phi(X)$ for the given sample set $X$. The embedding is then fed along with sample set to TAN model parameterized by $W_2$. We then optimize the following modified objective:
\vspace{-2mm}
\begin{equation}
\min_{W_1,W_2} -\frac{1}{N}\sum_i \frac{1}{m_i} \sum_j \log p_{W_2}\left(x_{ij}|\phi_{W_1}(X_{i \setminus j} ) \right).
\vspace{-2mm}
\end{equation}
%
%As a proof-of-concept, we first perform a synthetic experiment where $\mathcal{P}_\theta=\mathcal{N}(\theta,1)$ with $\theta \in \mathbb{R}^2$. In the training data we provide $N=1000$ sample sets with $\theta$ uniformly distributed in $[-10,10]^2$ and 10\% of the $\theta$ being repeated. We test on a equally spaced grid over $[-20,20]^2$. Using DeepSets+TAN we could achieve log likelihood of -3.97 compared to true value of $-\log(2\pi e)=-2.83$.
%
%As a more complex task, we 
We attempt to model point-cloud representation of objects from ShapeNet~\citep{chang2015shapenet}. We produce point-clouds with 1000 particles each ($x,y,z$-coordinates) from the mesh representation of objects using the point-cloud-library's sampling routine~\cite{Rusu_ICRA2011_PCL}. We consider each category of objects (\eg aeroplane, chair, car) as a family and each point cloud for each object in the category as a sample set. We train a TAN and only show samples in \reffig{fig:pc_samples} produced for unseen test sets, as there are neither any baselines for this task nor ground truth for likelihood.
From the samples, we see that our model is able to capture the structure of different kinds of unseen aeroplanes and chairs, with very few artifacts in samples, which reflects a good fit.

Note that this task is subtly different from conditional density estimation as we do not have access to class/parameter values during training. Also we want to caution users against using this method when the test sample set is very different from training or comes from a different family distribution.

\vspace{-2mm}
\section{Conclusion}
\vspace{-1mm}
In this work, we showed that we can significantly improve density estimation for real valued data by jointly leveraging transformations of variables with autoregressive models and proposed novel modules for both. 
%We show a considerable improvement with our methods through a comprehensive study over both real world and synthetic data. Also, we illustrate the utility of our models in outlier detection and digit modeling tasks.
We systematically characterized various modules and evaluated their contributions in a comprehensive ablation study.
This exercise not only re-emphasized the benefits of joint modeling, but also revealed some straightforward modules and combinations thereof, which are empirically good, but were missed earlier, \eg the untied linear conditionals. Finally we introduced a novel data driven framework for learning a family of distributions.

\section*{Acknowledgements} This research is partly funded by DOE grant DESC0011114, NSF IIS1563887, NIH R01GM114311, NSF IIS1447676, and the DARPA D3M program.

\bibliographystyle{icml2018}
\bibliography{ref}

\clearpage
\appendix
\onecolumn
\section{Appendix}
Below we detail the results on several datasets using different combinations of transformations and autoregressive conditional models. Each additive coupling transformation uses a fully connected network with two hidden layers of 256 units. RNN transformations use a hidden state with 16 units. SingleInd conditional models modeled each dimension's conditional as a standard Gaussian. MultiInd modeled each dimension's conditional as independent mixtures with 40 components (each with mean, scale, and weight parameter). RAM, LAM, and Tied conditional models each had a hidden state with 120 units that was fed through two fully connected layers each with 120 units to produce the parameters of the mixtures with 40 components. The RAM hidden state was produced by a GRU with 256 units. LAM and Tied hidden states came through a linear map as discussed above.

\begin{table*}[ht!]
    \caption{Held out test log-likelihoods for the Markovian dataset. The superscripts denote rankings of log-likelihoods on the validation dataset. Note that NADE is TIED conditional with None Transform and NICE is Add+Re Transformation with SingleInd Conditional. In parenthesis is the top-10 picks using valiation set. \label{tab:markov}}
    \centering
{\tiny
\resizebox{1.0\textwidth}{!}{
\begin{tabular}{|l|c|c|c|c|c|} \hline
{\bfseries Transformation}&{\bfseries LAM}&{\bfseries RAM}&{\bfseries TIED}&{\bfseries MultiInd}&{\bfseries SingleInd}\\ \hline
{\tiny None}&$14.319$&$-29.950$&$-0.612$&$-41.472$&${---}$\\ \hline
{\tiny L None}&${\bm{ 15.486^{(9)}}}$&$14.538$&$10.906$&$5.252$&$-9.426$\\ \hline
{\tiny RNN}&$14.777$&$-37.716$&$11.075$&$-30.491$&$-37.038$\\ \hline
{\tiny L RNN}&${\bm{ 15.658^{(5)}}}$&$10.354$&$10.910$&$5.370$&$3.310$\\ \hline
{\tiny 2xRNN}&$14.683$&$13.698$&$11.493$&$-18.448$&$-34.268$\\ \hline
{\tiny L 2xRNN}&${\bm{ 15.474^{(8)}}}$&${\bm{ 15.752^{(3)}}}$&$12.316$&$5.385$&$3.739$\\ \hline
{\tiny 4xAdd+Re}&$15.269$&$12.257$&$12.912$&$12.446$&$11.625$\\ \hline
{\tiny L 4xAdd+Re}&${\bm{ 15.683^{(6)}}}$&$12.594$&$13.845$&$12.768$&$12.069$\\ \hline
{\tiny 4xSRNN+Re}&$14.829$&$14.381$&$11.798$&$11.738$&$12.932$\\ \hline
{\tiny L 4xSRNN+Re}&$15.289$&${\bm{ 16.202^{(1)}}}$&$12.748$&${\bm{ 15.415^{(10)}}}$&$13.908$\\ \hline
{\tiny RNN+4xAdd+Re}&$15.171$&$12.991$&$14.455$&$11.467$&$10.382$\\ \hline
{\tiny L RNN+4xAdd+Re}&$15.078$&$12.655$&$14.415$&$12.886$&$12.315$\\ \hline
{\tiny RNN+4xSRNN+Re}&$14.968$&${\bm{ 16.216^{(2)}}}$&$12.590$&${\bm{ 15.589^{(4)}}}$&$14.231$\\ \hline
{\tiny L RNN+4xSRNN+Re}&$15.429$&${\bm{ 15.566^{(7)}}}$&$14.179$&$14.528$&$13.961$\\ \hline
\end{tabular}}}
\end{table*}

{
\begin{table*}[ht!]
 \caption{Held out test log-likelihoods for star 32d dataset. The superscript denotes ranking of log-likelihood on cross validation dataset. Note that NADE is TIED conditional with None Transform and NICE is Add+Re Transformation with SingleInd Conditional. In parenthesis is the top-10 picks using valiation set.\label{tab:star}}
    \centering
    {\tiny
\resizebox{1.0\textwidth}{!}{
\begin{tabular}{|l|c|c|c|c|c|} \hline
{\bfseries Transformation}&{\bfseries LAM}&{\bfseries RAM}&{\bfseries TIED}&{\bfseries MultiInd}&{\bfseries SingleInd}\\ \hline
{\tiny None}&$-2.041$&$2.554$&$-10.454$&$-29.485$&${---}$\\ \hline
{\tiny L None}&$5.454$&$8.247$&$-7.858$&$-26.988$&$-38.952$\\ \hline
{\tiny RNN}&$-1.276$&$2.762$&$-6.292$&$-25.946$&$-41.275$\\ \hline
{\tiny L RNN}&$7.775$&$6.335$&$-1.157$&$-25.986$&$-34.408$\\ \hline
{\tiny 2xRNN}&$3.705$&$8.032$&$-0.565$&$-25.100$&$-38.490$\\ \hline
{\tiny L 2xRNN}&${\bm{ 14.878^{(3)}}}$&$9.946$&$0.901$&$-23.772$&$-33.075$\\ \hline
{\tiny 4xAdd+Re}&${\bm{ 13.278^{(6)}}}$&${\bm{ 11.561^{(9)}}}$&$7.146$&$-16.740$&$-21.332$\\ \hline
{\tiny L 4xAdd+Re}&${\bm{ 15.728^{(2)}}}$&${\bm{ 12.444^{(7)}}}$&$9.031$&$-6.091$&$-11.225$\\ \hline
{\tiny 4xSRNN+Re}&$3.496$&$8.429$&$-1.380$&$-15.590$&$-23.712$\\ \hline
{\tiny L 4xSRNN+Re}&${\bm{ 16.042^{(1)}}}$&${\bm{ 9.939^{(10)}}}$&$5.598$&$-12.530$&$-16.889$\\ \hline
{\tiny RNN+4xAdd+Re}&${\bm{ 14.071^{(5)}}}$&${\bm{ 14.123^{(4)}}}$&$6.868$&$-14.773$&$-20.483$\\ \hline
{\tiny L RNN+4xAdd+Re}&${\bm{ 11.819^{(8)}}}$&$9.253$&$2.638$&$-7.662$&$-14.530$\\ \hline
{\tiny RNN+4xSRNN+Re}&$-0.679$&$3.320$&$-6.172$&$-12.879$&$-19.204$\\ \hline
{\tiny L RNN+4xSRNN+Re}&$7.433$&$7.324$&$3.554$&$-10.427$&$-15.243$\\ \hline
\end{tabular}}}
\end{table*}
}

{
\begin{table*}[ht]
\caption{Held out test log-likelihood for Star 128d dataset.The superscript denotes ranking of log-likelihood on crossvalidation dataset. Note that NADE is TIED conditional with None Transform and NICE is Add+Re Transformation with SingleInd Conditional. In parenthesis is the top-10 picks using valiation set.\label{tab:star128}}
\centering
{\tiny 
\resizebox{1.0\textwidth}{!}{
\begin{tabular}{|l|c|c|c|c|c|} \hline
{\bfseries Transformation}&{\bfseries LAM}&{\bfseries RAM}&{\bfseries TIED}&{\bfseries MultiInd}&{\bfseries SingleInd}\\ \hline
{\tiny None}&$15.671$&$15.895$&$-83.115$&$-128.238$&${---}$\\ \hline
{\tiny L None}&$57.881$&$-82.100$&$-28.206$&$-123.939$&$-159.391$\\ \hline
{\tiny RNN}&$18.766$&$48.295$&$-22.485$&$-113.181$&$-178.641$\\ \hline
{\tiny L RNN}&${\bm{ 66.070^{(9)}}}$&$-49.084$&$31.136$&$-107.083$&$-155.324$\\ \hline
{\tiny 2xRNN}&$27.295$&$45.834$&$-11.930$&$-113.210$&$-178.331$\\ \hline
{\tiny L 2xRNN}&${\bm{ 85.681^{(3)}}}$&$-84.524$&$30.974$&$-105.368$&$-162.635$\\ \hline
{\tiny 4xAdd+Re}&${\bm{ 77.195^{(6)}}}$&${\bm{ 61.947^{(10)}}}$&$16.062$&$-75.206$&$-111.542$\\ \hline
{\tiny L 4xAdd+Re}&${\bm{ 88.837^{(1)}}}$&$-21.882$&$20.234$&$-65.694$&$-96.071$\\ \hline
{\tiny 4xSRNN+Re}&$33.577$&$-98.796$&$3.256$&$-88.912$&$-98.936$\\ \hline
{\tiny L 4xSRNN+Re}&${\bm{ 86.375^{(2)}}}$&${\bm{ 76.968^{(5)}}}$&$33.481$&$-85.590$&$-93.086$\\ \hline
{\tiny RNN+4xAdd+Re}&${\bm{ 66.540^{(8)}}}$&$-57.861$&$-16.277$&$-75.491$&$-114.729$\\ \hline
{\tiny L RNN+4xAdd+Re}&${\bm{ 80.063^{(4)}}}$&$32.104$&$21.944$&$-71.933$&$-100.384$\\ \hline
{\tiny RNN+4xSRNN+Re}&$21.719$&$-87.335$&$-6.517$&$-76.459$&$-85.422$\\ \hline
{\tiny L RNN+4xSRNN+Re}&${\bm{ 72.463^{(7)}}}$&$56.201$&$26.269$&$-71.843$&$-91.695$\\ \hline
\end{tabular}}}
\end{table*}
}

{
\begin{table*}
\caption{Average performance percentage score for each model across all datasets. Note that this measure is not over a logarithmic space. \label{tab:combined_scored}}
    \centering
    {\tiny
\resizebox{1.0\textwidth}{!}{
\begin{tabular}{|l|c|c|c|c|c||c|} \hline
{\bfseries Transformation}&{\bfseries LAM}&{\bfseries RAM}&{\bfseries TIED}&{\bfseries MultiInd}&{\bfseries SingleInd} & MAX \\ \hline
None&$0.218$ &$0.118$ &$0.006$ &$0.000$ &$0.000$ &$0.218$\\ \hline
L None&$0.154$ &$0.179$ &$0.026$ &$0.051$ &$0.001$ &$0.179$\\ \hline
RNN&$0.086$ &$0.158$ &$0.014$ &$0.001$ &$0.000$ &$0.158$\\ \hline
L RNN&$0.173$ &{ $\bm{0.540}$} &$0.014$ &$0.040$ &$0.013$ &$0.540$\\ \hline
2xRNN&$0.151$ &$0.101$ &$0.045$ &$0.001$ &$0.000$ &$0.151$\\ \hline
L 2xRNN&$0.118$ &$0.330$ &$0.015$ &$0.045$ &$0.025$ &$0.330$\\ \hline
4xAdd+Re&$0.036$ &$0.047$ &$0.015$ &$0.010$ &$0.006$ &$0.047$\\ \hline
L 4xAdd+Re&$0.153$ &$0.096$ &$0.025$ &$0.014$ &$0.009$ &$0.153$\\ \hline
4xSRNN+Re&$0.086$ &$0.051$ &$0.031$ &$0.010$ &$0.008$ &$0.086$\\ \hline
L 4xSRNN+Re&$0.109$ &$0.143$ &$0.023$ &$0.021$ &$0.018$ &$0.143$\\ \hline
RNN+4xAdd+Re&$0.121$ &$0.096$ &$0.023$ &$0.011$ &$0.011$ &$0.121$\\ \hline
L RNN+4xAdd+Re&$0.336$ &$0.165$ &$0.024$ &$0.016$ &$0.013$ &$0.336$\\ \hline
RNN+4xSRNN+Re&$0.102$ &$0.151$ &$0.017$ &$0.012$ &$0.014$ &$0.151$\\ \hline
L RNN+4xSRNN+Re&$0.211$ &$0.288$ &$0.024$ &$0.018$ &$0.016$ &$0.288$\\ \hline \hline
MAX &$0.336$ &$0.540$ &$0.045$ &$0.051$ &$0.025$ & \\ \hline
\end{tabular}}}
\end{table*}
}

{
\begin{table*}
\caption{Held out test log-likelihood for \texttt{forest} dataset.The superscript denotes ranking of log-likelihood on crossvalidation dataset. Note that NADE is TIED conditional with None Transform and NICE is Add+Re Transformation with SingleInd Conditional. In parenthesis is the top-10 picks using valiation set.\label{tab:forestllk}}
\centering
{\tiny 
\resizebox{1.0\textwidth}{!}{
\begin{tabular}{|l|c|c|c|c|c|} \hline
{\bfseries Transformation}&{\bfseries LAM}&{\bfseries RAM}&{\bfseries TIED}&{\bfseries MultiInd}&{\bfseries SingleInd}\\ \hline
{\tiny None}&$0.751$&$-1.383$&$-0.653$&$-12.824$&${---}$\\ \hline
{\tiny L None}&$1.910$&$1.834$&$-0.243$&$-7.665$&$-11.062$\\ \hline
{\tiny RNN}&$1.395$&$0.053$&$0.221$&$-5.130$&$-15.983$\\ \hline
{\tiny L RNN}&${\bm{ 2.189^{(8)}}}$&$1.747$&$-0.087$&$-4.001$&$-5.807$\\ \hline
{\tiny 2xRNN}&$1.832$&$1.830$&$0.448$&$-6.162$&$-9.095$\\ \hline
{\tiny L 2xRNN}&${\bm{ 2.240^{(6)}}}$&${\bm{ 2.432^{(3)}}}$&$0.264$&$-3.956$&$-5.125$\\ \hline
{\tiny 4xAdd+Re}&$1.106$&$1.430$&$0.420$&$-0.021$&$-0.492$\\ \hline
{\tiny L 4xAdd+Re}&$2.043$&$1.979$&$0.909$&$0.365$&$-0.088$\\ \hline
{\tiny 4xSRNN+Re}&$1.178$&$1.428$&$0.187$&$-0.029$&$-0.212$\\ \hline
{\tiny L 4xSRNN+Re}&${\bm{ 2.089^{(9)}}}$&${\bm{ 2.061^{(10)}}}$&$0.611$&$0.754$&$0.593$\\ \hline
{\tiny RNN+4xAdd+Re}&$1.962$&${\bm{ 2.226^{(7)}}}$&$0.857$&$0.081$&$0.086$\\ \hline
{\tiny L RNN+4xAdd+Re}&${\bm{ 2.389^{(4)}}}$&${\bm{ 2.672^{(1)}}}$&$0.852$&$0.450$&$0.251$\\ \hline
{\tiny RNN+4xSRNN+Re}&$1.599$&$1.545$&$0.510$&$0.182$&$0.369$\\ \hline
{\tiny L RNN+4xSRNN+Re}&${\bm{ 2.297^{(5)}}}$&${\bm{ 2.443^{(2)}}}$&$0.804$&$0.600$&$0.480$\\ \hline
\end{tabular}}}
\end{table*}
}

{
\begin{table*}
\caption{Held out test log-likelihood for \texttt{pendigits} dataset. The superscript denotes ranking of log-likelihood on crossvalidation dataset. Note that NADE is TIED conditional with None Transform and NICE is Add+Re Transformation with SingleInd Conditional. In parenthesis is the top-10 picks using valiation set.\label{tab:pendigits}}
\centering
{\tiny 
\resizebox{1.0\textwidth}{!}{
\begin{tabular}{|l|c|c|c|c|c|} \hline
{\bfseries Transformation}&{\bfseries LAM}&{\bfseries RAM}&{\bfseries TIED}&{\bfseries MultiInd}&{\bfseries SingleInd}\\ \hline
{\tiny None}&${\bm{ 6.923^{(1)}}}$&${\bm{ 3.911^{(8)}}}$&$1.437$&$-14.138$&${---}$\\ \hline
{\tiny L None}&${\bm{ 4.104^{(9)}}}$&$2.911$&$-2.872$&$-9.997$&$-15.617$\\ \hline
{\tiny RNN}&${\bm{ 5.464^{(3)}}}$&$3.273$&$-1.676$&$-10.144$&$-19.719$\\ \hline
{\tiny L RNN}&${\bm{ 4.072^{(6)}}}$&$1.398$&$-2.299$&$-10.840$&$-13.103$\\ \hline
{\tiny 2xRNN}&${\bm{ 6.376^{(5)}}}$&${\bm{ 3.896^{(7)}}}$&$-4.002$&$-12.132$&$-16.576$\\ \hline
{\tiny L 2xRNN}&$2.987$&$0.871$&$-3.977$&$-10.890$&$-12.711$\\ \hline
{\tiny 4xAdd+Re}&$-1.924$&$-3.087$&$-3.172$&$-5.010$&$-6.498$\\ \hline
{\tiny L 4xAdd+Re}&$-1.796$&$-1.438$&$-2.288$&$-4.951$&$-7.834$\\ \hline
{\tiny 4xSRNN+Re}&${\bm{ 5.854^{(2)}}}$&$2.146$&$-2.827$&$-5.970$&$-7.084$\\ \hline
{\tiny L 4xSRNN+Re}&$3.758$&$-1.020$&$-3.370$&$-5.885$&$-12.978$\\ \hline
{\tiny RNN+4xAdd+Re}&$-2.357$&$-2.869$&$-2.187$&$-5.454$&$-8.053$\\ \hline
{\tiny L RNN+4xAdd+Re}&$-2.687$&$-2.103$&$-2.185$&$-4.742$&$-6.941$\\ \hline
{\tiny RNN+4xSRNN+Re}&${\bm{ 5.207^{(4)}}}$&$2.425$&$-2.126$&$-5.147$&$-8.859$\\ \hline
{\tiny L RNN+4xSRNN+Re}&${\bm{ 3.466^{(10)}}}$&$0.496$&$-2.761$&$-7.205$&$-13.897$\\ \hline
\end{tabular}}}
\end{table*}
}

{
\begin{table*}
\caption{Held out test log-likelihood for \texttt{susy} dataset.The superscript denotes ranking of log-likelihood on crossvalidation dataset. Note that NADE is TIED conditional with None Transform and NICE is Add+Re Transformation with SingleInd Conditional. In parenthesis is the top-10 picks using valiation set.\label{tab:susy}}
\centering
{\tiny 
\resizebox{1.0\textwidth}{!}{
\begin{tabular}{|l|c|c|c|c|c|} \hline
{\bfseries Transformation}&{\bfseries LAM}&{\bfseries RAM}&{\bfseries TIED}&{\bfseries MultiInd}&{\bfseries SingleInd}\\ \hline
{\tiny None}&$9.736$&$-14.821$&$-5.721$&$-21.369$&${---}$\\ \hline
{\tiny L None}&$15.731$&${\bm{ 16.930^{(8)}}}$&$6.410$&$-8.846$&$-17.130$\\ \hline
{\tiny RNN}&$12.784$&$3.347$&$6.114$&$-18.575$&$-44.273$\\ \hline
{\tiny L RNN}&$16.381$&${\bm{ 18.389^{(2)}}}$&$6.772$&$-5.744$&$-11.489$\\ \hline
{\tiny 2xRNN}&$11.052$&$14.362$&$3.595$&$-16.478$&$-33.126$\\ \hline
{\tiny L 2xRNN}&$14.523$&${\bm{ 17.373^{(7)}}}$&$10.687$&$-6.884$&$-10.420$\\ \hline
{\tiny 4xAdd+Re}&$9.835$&$8.033$&$7.238$&$6.031$&$4.245$\\ \hline
{\tiny L 4xAdd+Re}&${\bm{ 17.673^{(3)}}}$&${\bm{ 16.500^{(10)}}}$&$11.613$&$10.941$&$9.034$\\ \hline
{\tiny 4xSRNN+Re}&$8.798$&$13.235$&$1.234$&$6.936$&$3.378$\\ \hline
{\tiny L 4xSRNN+Re}&$14.242$&${\bm{ 17.870^{(5)}}}$&$15.397$&$12.161$&$13.413$\\ \hline
{\tiny RNN+4xAdd+Re}&$15.408$&$12.480$&$9.409$&$7.619$&$5.446$\\ \hline
{\tiny L RNN+4xAdd+Re}&${\bm{ 17.474^{(6)}}}$&$16.376$&$13.765$&$10.951$&$8.269$\\ \hline
{\tiny RNN+4xSRNN+Re}&$14.066$&${\bm{ 17.691^{(4)}}}$&$9.136$&$10.088$&$7.656$\\ \hline
{\tiny L RNN+4xSRNN+Re}&${\bm{ 16.627^{(9)}}}$&${\bm{ 18.941^{(1)}}}$&$13.469$&$12.105$&$12.349$\\ \hline
\end{tabular}}}
\end{table*}
}

{
\begin{table*}
\caption{Held out test log-likelihood for \texttt{higgs} dataset.The superscript denotes ranking of log-likelihood on crossvalidation dataset. Note that NADE is TIED conditional with None Transform and NICE is Add+Re Transformation with SingleInd Conditional. In parenthesis is the top-10 picks using valiation set.\label{tab:higgs}}
\centering
{\tiny 
\resizebox{1.0\textwidth}{!}{
\begin{tabular}{|l|c|c|c|c|c|} \hline
{\bfseries Transformation}&{\bfseries LAM}&{\bfseries RAM}&{\bfseries TIED}&{\bfseries MultiInd}&{\bfseries SingleInd}\\ \hline
{\tiny None}&$-6.220$&$-5.848$&$-13.883$&$-25.793$&${---}$\\ \hline
{\tiny L None}&${\bm{ -3.798^{(8)}}}$&$-10.651$&$-9.084$&$-16.025$&$-36.051$\\ \hline
{\tiny RNN}&$-5.800$&${\bm{ -2.600^{(3)}}}$&$-10.797$&$-25.760$&$-66.223$\\ \hline
{\tiny L RNN}&${\bm{ -3.975^{(9)}}}$&${\bm{ -0.340^{(1)}}}$&$-8.574$&$-18.607$&$-32.753$\\ \hline
{\tiny 2xRNN}&$-6.456$&$-4.833$&$-9.192$&$-25.398$&$-60.040$\\ \hline
{\tiny L 2xRNN}&$-5.866$&${\bm{ -3.222^{(5)}}}$&$-8.216$&$-16.083$&$-30.730$\\ \hline
{\tiny 4xAdd+Re}&$-6.502$&$-10.491$&$-9.356$&$-13.678$&$-15.138$\\ \hline
{\tiny L 4xAdd+Re}&$-5.377$&$-5.611$&$-8.006$&$-12.106$&$-14.129$\\ \hline
{\tiny 4xSRNN+Re}&$-7.422$&$-6.863$&$-11.033$&$-11.878$&$-12.182$\\ \hline
{\tiny L 4xSRNN+Re}&$-5.999$&$-9.329$&$-8.474$&$-8.223$&$-8.926$\\ \hline
{\tiny RNN+4xAdd+Re}&${\bm{ -4.242^{(10)}}}$&$-4.804$&$-9.187$&$-12.321$&$-15.261$\\ \hline
{\tiny L RNN+4xAdd+Re}&${\bm{ -3.396^{(6)}}}$&${\bm{ -3.049^{(4)}}}$&$-8.052$&$-12.246$&$-13.765$\\ \hline
{\tiny RNN+4xSRNN+Re}&$-5.262$&${\bm{ -2.116^{(2)}}}$&$-10.105$&$-12.307$&$-9.388$\\ \hline
{\tiny L RNN+4xSRNN+Re}&${\bm{ -3.756^{(7)}}}$&$-4.773$&$-8.097$&$-9.378$&$-7.721$\\ \hline
\end{tabular}}}
\end{table*}
}

{
\begin{table*}
\caption{Held out test log-likelihood for \texttt{hepmass} dataset.The superscript denotes ranking of log-likelihood on crossvalidation dataset. Note that NADE is TIED conditional with None Transform and NICE is Add+Re Transformation with SingleInd Conditional. In parenthesis is the top-10 picks using valiation set.\label{tab:hepmass}}
\centering
{\tiny 
\resizebox{1.0\textwidth}{!}{
\begin{tabular}{|l|c|c|c|c|c|} \hline
{\bfseries Transformation}&{\bfseries LAM}&{\bfseries RAM}&{\bfseries TIED}&{\bfseries MultiInd}&{\bfseries SingleInd}\\ \hline
{\tiny None}&$2.328$&${\bm{ 3.710^{(6)}}}$&$-4.948$&$-19.771$&$---$\\ \hline
{\tiny L None}&${\bm{ 3.570^{(7)}}}$&$2.517$&$-4.052$&$-9.266$&$-35.042$\\ \hline
{\tiny RNN}&$2.088$&${\bm{ 4.935^{(1)}}}$&$-1.639$&$-19.851$&$-47.686$\\ \hline
{\tiny L RNN}&${\bm{ 2.869^{(10)}}}$&${\bm{ 5.047^{(2)}}}$&$-2.920$&$-16.032$&$-30.210$\\ \hline
{\tiny 2xRNN}&$1.774$&$0.902$&$-1.909$&$-15.440$&$-36.754$\\ \hline
{\tiny L 2xRNN}&$2.053$&${\bm{ 3.680^{(5)}}}$&$-2.150$&$-15.457$&$-24.079$\\ \hline
{\tiny 4xAdd+Re}&$1.678$&$1.873$&$-4.046$&$-9.117$&$-11.387$\\ \hline
{\tiny L 4xAdd+Re}&$1.961$&$2.543$&$-2.259$&$-6.907$&$-9.275$\\ \hline
{\tiny 4xSRNN+Re}&$1.443$&$2.156$&$-2.904$&$-6.091$&$-7.186$\\ \hline
{\tiny L 4xSRNN+Re}&$2.072$&$2.730$&$-3.014$&$-5.747$&$-6.245$\\ \hline
{\tiny RNN+4xAdd+Re}&$2.817$&$0.912$&$-2.514$&$-6.003$&$-9.284$\\ \hline
{\tiny L RNN+4xAdd+Re}&${\bm{ 3.906^{(3)}}}$&$-1.869$&$-3.847$&$-6.339$&$-9.103$\\ \hline
{\tiny RNN+4xSRNN+Re}&$2.663$&${\bm{ 3.586^{(8)}}}$&$-0.863$&$-7.146$&$-3.939$\\ \hline
{\tiny L RNN+4xSRNN+Re}&${\bm{ 3.759^{(4)}}}$&${\bm{ 3.487^{(9)}}}$&$-0.239$&$-7.522$&$-6.102$\\ \hline
\end{tabular}}}
\end{table*}
}

{

\begin{table*}
\caption{Held out test log-likelihood for \texttt{satimage2} dataset.The superscript denotes ranking of log-likelihood on crossvalidation dataset. Note that NADE is TIED conditional with None Transform and NICE is Add+Re Transformation with SingleInd Conditional. In parenthesis is the top-10 picks using valiation set.\label{tab:satimage2}}
\centering
{\tiny 
\resizebox{1.0\textwidth}{!}{
\begin{tabular}{|l|c|c|c|c|c|} \hline
{\bfseries Transformation}&{\bfseries LAM}&{\bfseries RAM}&{\bfseries TIED}&{\bfseries MultiInd}&{\bfseries SingleInd}\\ \hline
{\tiny None}&${\bm{ -1.716^{(9)}}}$&${\bm{ -1.257^{(3)}}}$&$-9.296$&$-50.507$&${---}$\\ \hline
{\tiny L None}&$-20.164$&${\bm{ -1.079^{(4)}}}$&$-2.635$&${\bm{ -1.570^{(5)}}}$&$-5.972$\\ \hline
{\tiny RNN}&$-7.728$&$-4.949$&$-5.466$&$-6.047$&$-16.521$\\ \hline
{\tiny L RNN}&$-31.296$&${\bm{ -0.773^{(2)}}}$&$-3.944$&${\bm{ -1.824^{(8)}}}$&$-2.977$\\ \hline
{\tiny 2xRNN}&$-12.283$&${\bm{ -2.193^{(7)}}}$&$-2.137$&$-5.447$&$-8.075$\\ \hline
{\tiny L 2xRNN}&$-20.968$&${\bm{ -0.550^{(1)}}}$&$-5.140$&${\bm{ -1.699^{(6)}}}$&${\bm{ -2.276^{(10)}}}$\\ \hline
{\tiny 4xAdd+Re}&$-19.931$&$-7.539$&$-11.826$&$-18.901$&$-17.977$\\ \hline
{\tiny L 4xAdd+Re}&$-21.128$&$-9.944$&$-12.336$&$-21.677$&$-24.070$\\ \hline
{\tiny 4xSRNN+Re}&$-7.519$&$-11.368$&$-2.549$&$-7.730$&$-7.232$\\ \hline
{\tiny L 4xSRNN+Re}&$-18.170$&$-7.709$&$-5.533$&$-17.085$&$-15.347$\\ \hline
{\tiny RNN+4xAdd+Re}&$-19.278$&$-11.789$&$-12.837$&$-21.249$&$-22.786$\\ \hline
{\tiny L RNN+4xAdd+Re}&$-20.899$&$-12.949$&$-12.867$&$-26.164$&$-28.302$\\ \hline
{\tiny RNN+4xSRNN+Re}&$-13.476$&$-3.951$&$-6.284$&$-15.025$&$-16.443$\\ \hline
{\tiny L RNN+4xSRNN+Re}&$-20.179$&$-12.128$&$-7.258$&$-18.065$&$-18.125$\\ \hline
\end{tabular}}}
\end{table*}
}

{
\begin{table*}
\caption{Held out test log-likelihood for \texttt{music} dataset.The superscript denotes ranking of log-likelihood on crossvalidation dataset. Note that NADE is TIED conditional with None Transform and NICE is Add+Re Transformation with SingleInd Conditional. In parenthesis is the top-10 picks using valiation set.\label{tab:music}}
\centering
{\tiny 
\resizebox{1.0\textwidth}{!}{
\begin{tabular}{|l|c|c|c|c|c|} \hline
{\bfseries Transformation}&{\bfseries LAM}&{\bfseries RAM}&{\bfseries TIED}&{\bfseries MultiInd}&{\bfseries SingleInd}\\ \hline
{\tiny None}&$-57.873$&$-97.925$&$-98.047$&$-113.099$&${---}$\\ \hline
{\tiny L None}&${\bm{ -52.954^{(4)}}}$&$-74.220$&$-72.441$&$-82.866$&$-104.287$\\ \hline
{\tiny RNN}&${\bm{ -54.933^{(10)}}}$&$-80.436$&$-74.361$&$-106.219$&$-144.735$\\ \hline
{\tiny L RNN}&${\bm{ -52.710^{(3)}}}$&$-59.815$&$-66.536$&$-82.731$&$-98.813$\\ \hline
{\tiny 2xRNN}&$-56.958$&$-85.359$&$-77.456$&$-104.440$&$-133.898$\\ \hline
{\tiny L 2xRNN}&${\bm{ -53.956^{(8)}}}$&$-57.611$&$-65.016$&$-82.678$&$-96.542$\\ \hline
{\tiny 4xAdd+Re}&$-56.349$&$-69.302$&$-67.064$&$-73.886$&$-83.524$\\ \hline
{\tiny L 4xAdd+Re}&${\bm{ -53.169^{(5)}}}$&$-59.282$&$-59.093$&$-69.887$&$-79.330$\\ \hline
{\tiny 4xSRNN+Re}&$-57.670$&$-68.116$&$-74.006$&$-78.032$&$-121.197$\\ \hline
{\tiny L 4xSRNN+Re}&${\bm{ -53.879^{(7)}}}$&$-55.665$&$-63.894$&$-77.564$&$-81.188$\\ \hline
{\tiny RNN+4xAdd+Re}&${\bm{ -53.177^{(6)}}}$&$-67.377$&$-63.372$&$-73.882$&$-84.032$\\ \hline
{\tiny L RNN+4xAdd+Re}&${\bm{ -51.572^{(1)}}}$&$-56.190$&$-58.885$&$-69.484$&$-79.555$\\ \hline
{\tiny RNN+4xSRNN+Re}&${\bm{ -54.065^{(9)}}}$&$-61.204$&$-76.437$&$-71.814$&$-81.087$\\ \hline
{\tiny L RNN+4xSRNN+Re}&${\bm{ -52.617^{(2)}}}$&$-68.756$&$-65.061$&$-83.292$&$-78.997$\\ \hline
\end{tabular}}}
\end{table*}
}

{
\begin{table*}
\caption{Held out test log-likelihood for \texttt{wordvecs} dataset.The superscript denotes ranking of log-likelihood on validation dataset. Due to time constraints only models with linear transformations were trained. \label{tab:word2vec}}
\centering
{\tiny 
\resizebox{1.0\textwidth}{!}{
\begin{tabular}{|l|c|c|c|c|c|} \hline
{\bfseries Transformation}&{\bfseries LAM}&{\bfseries RAM}&{\bfseries TIED}&{\bfseries MultiInd}&{\bfseries SingleInd}\\ \hline
%{\tiny None}&$0.000$&$0.000$&$0.000$&$0.000$&${---}$\\ \hline
{\tiny L None}&${\bm{ -252.659^{(6)}}}$&$-279.788$&$-278.789$&$-332.474$&$-387.341$\\ \hline
%{\tiny RNN}&$0.000$&$0.000$&$0.000$&$0.000$&$0.000$\\ \hline
{\tiny L RNN}&${\bm{ -252.894^{(7)}}}$&$-278.795$&$-278.663$&$-332.689$&$-386.700$\\ \hline
%{\tiny 2xRNN}&$0.000$&$0.000$&$0.000$&$0.000$&$0.000$\\ \hline
{\tiny L 2xRNN}&${\bm{ -250.285^{(4)}}}$&$-275.508$&$-277.848$&$-333.234$&$-386.649$\\ \hline
%{\tiny 4xAdd+Re}&$0.000$&$0.000$&$0.000$&$0.000$&$0.000$\\ \hline
{\tiny L 4xAdd+Re}&${\bm{ -247.440^{(1)}}}$&${\bm{ -272.371^{(8)}}}$&$-274.205$&$-331.148$&$-374.563$\\ \hline
%{\tiny 4xSRNN+Re}&$0.000$&$-353.165$&$0.000$&$0.000$&$0.000$\\ \hline
{\tiny L 4xSRNN+Re}&${\bm{ -248.393^{(2)}}}$&$-300.666$&${\bm{ -273.372^{(9)}}}$&$-308.735$&$0.000$\\ \hline
%{\tiny RNN+4xAdd+Re}&$0.000$&$0.000$&$0.000$&$0.000$&$0.000$\\ \hline
{\tiny L RNN+4xAdd+Re}&${\bm{ -249.980^{(3)}}}$&$-280.938$&${\bm{ -273.976^{(10)}}}$&$-331.316$&$-380.031$\\ \hline
%{\tiny RNN+4xSRNN+Re}&$0.000$&$0.000$&$0.000$&$0.000$&$0.000$\\ \hline
{\tiny L RNN+4xSRNN+Re}&${\bm{ -251.468^{(5)}}}$&$-280.325$&$-274.082$&$-308.148$&$-395.084$\\ \hline
\end{tabular}}}
\end{table*}}

\end{document}